\documentclass[letterpaper]{article} % DO NOT CHANGE THIS
\usepackage{aaai2026}  % DO NOT CHANGE THIS
\usepackage{times}  % DO NOT CHANGE THIS
\usepackage{helvet}  % DO NOT CHANGE THIS
\usepackage{courier}  % DO NOT CHANGE THIS
\usepackage[hyphens]{url}  % DO NOT CHANGE THIS
\usepackage{graphicx} % DO NOT CHANGE THIS

\usepackage{natbib}  % DO NOT CHANGE THIS AND DO NOT ADD ANY OPTIONS TO IT
\usepackage{caption} % DO NOT CHANGE THIS AND DO NOT ADD ANY OPTIONS TO IT
\usepackage{array}
\usepackage{textcomp}
\usepackage{verbatim}

\usepackage{amsmath,amssymb,enumerate}
\usepackage{amsfonts}
\usepackage{mathtools}
\usepackage[mathscr]{euscript}
\usepackage{mathrsfs}
\usepackage[usenames,dvipsnames,svgnames,table, xcdraw]{xcolor} %http://ctan.org/pkg/xcolor
\usepackage[caption=false,font=footnotesize]{subfig}
\usepackage{booktabs}
\usepackage{multirow}
\usepackage[ruled,vlined,linesnumbered]{algorithm2e}
\SetKw{Continue}{continue}
\SetKw{Break}{break}

\usepackage{enumitem}
\usepackage{amsthm}

\usepackage{cite} % to group references
\usepackage[utf8]{inputenc}

\usepackage{pifont}

\usepackage{placeins}
\usepackage{tcolorbox}
\tcbuselibrary{listingsutf8}
\usepackage{inconsolata} % Optional: better monospace font

\definecolor{lightgray}{gray}{0.95}
\definecolor{darkgreen}{rgb}{0,0.5,0}
\definecolor{darkblue}{rgb}{0.0, 0.0, 0.5}
\definecolor{darkred}{rgb}{0.5, 0.0, 0.0}

\lstdefinelanguage{PDDL}{
    morekeywords={define,domain,problem,action,predicates,parameters,precondition,effect,and,not},
    sensitive=true,
    morecomment=[l]{;},
}

\lstdefinestyle{pddl}{
    language=PDDL,
    basicstyle=\ttfamily\scriptsize,
    keywordstyle=\color{darkblue}\bfseries,
    commentstyle=\color{darkgreen},
    stringstyle=\color{darkred},
    showstringspaces=false,
    breaklines=true,
    tabsize=2
}

\newif\ifshowcomment
\newif\ifnumberrevision
\newif\ifcolorrevision
\newif\ifstrikeremovel
\newcommand{\Ical}{\mathcal{I}}
\newcommand{\Jcal}{\mathcal{J}}

\newcommand{\move}{\textit{move}}
\newcommand{\wait}{\textit{wait}}
\newcommand{\complete}{\textit{complete}}

\newcommand{\lacam}{\textit{LaCAM}}
\newcommand{\brlacam}{\textit{BR-LaCAM}}
\newcommand{\brpibt}{\textit{BR-PIBT}}

\newcommand{\problemacronymP}{BRaP}

\newcommand{\problemupper}{Block Rearrangement}
\newcommand{\problemupperP}{Block Rearrangement Problem}
\newcommand{\gridU}{Grid}
\newcommand{\gridL}{grid}
\newcommand{\blockU}{Block}
\newcommand{\blockL}{block}

\newcommand{\tabref}{Table}
\newcommand{\figref}{Figure}
\newcommand{\algof}{Algorithm}

\SetCommentSty{mycommfont}

\newcommand{\llIf}[2]{{\DontPrintSemicolon\let\par\relax\lIf{#1}{#2}}}
\newcommand{\llElse}[1]{{\DontPrintSemicolon\let\par\relax\lElse{#1}}}

\title{Symbolic Planning and Multi-Agent Path Finding in Extremely Dense Environments with Unassigned Agents}
\author{
    \large
    Bo Fu\equalcontrib, Zhe Chen\equalcontrib, Rahul Chandan, Alexandre Ormiga Galvao Barbosa,\\ Michael Caldara, Joey W. Durham, and Federico Pecora
}
\affiliations{
    Amazon Robotics\\
    300 Riverpark Dr, North Reading, MA 01864 \\
    \{bofu, zhecm, rcd, aormiga, caldaram, josepdur, fpecora\}@amazon.com
}

\begin{document}

\maketitle

\begin{abstract}
We introduce the \problemupperP{} (\problemacronymP{}), a challenging component of large warehouse management which involves rearranging storage \blockL{}s within dense \gridL{}s to achieve a goal state. We formally define the \problemacronymP{} as a graph search problem. Building on intuitions from sliding puzzle problems, we propose five search-based solution algorithms, leveraging joint configuration space search, classical planning, multi-agent pathfinding, and expert heuristics. We evaluate the five approaches empirically for plan quality and scalability. Despite the exponential relation between search space size and \blockL{} number, our methods demonstrate efficiency in creating rearrangement plans for deeply buried \blockL{}s in up to 80\(\times\)80 \gridL{}s.
\end{abstract}

% \begin{IEEEkeywords}
% Multi-agent path finding, classical planning, PDDL, combinatorial optimization, sliding-tile puzzle
% \end{IEEEkeywords}

\section{Introduction}\label{sec:introduction}

\begin{figure}[t!]
    \centering
    \subfloat[Photo\label{fig:fulfillment_center_photo}]{
    \includegraphics[width=0.9\linewidth, trim=0 0 0 0, clip]{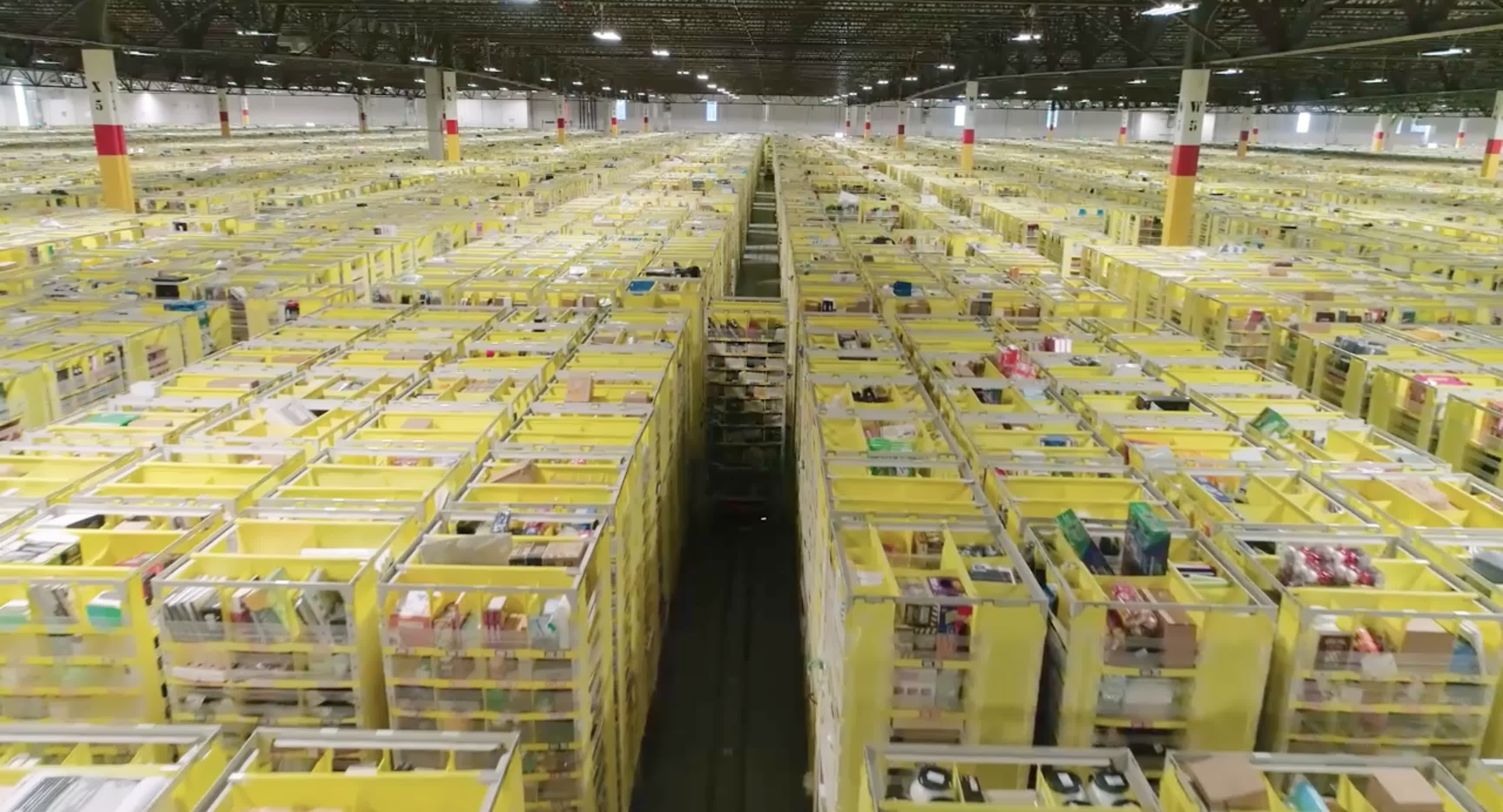}}
    \hfill
    \subfloat[Illustration diagram\label{fig:fulfillment_center_illustration}]{
    \includegraphics[width=0.95\linewidth, trim=0 0 0 0, clip]{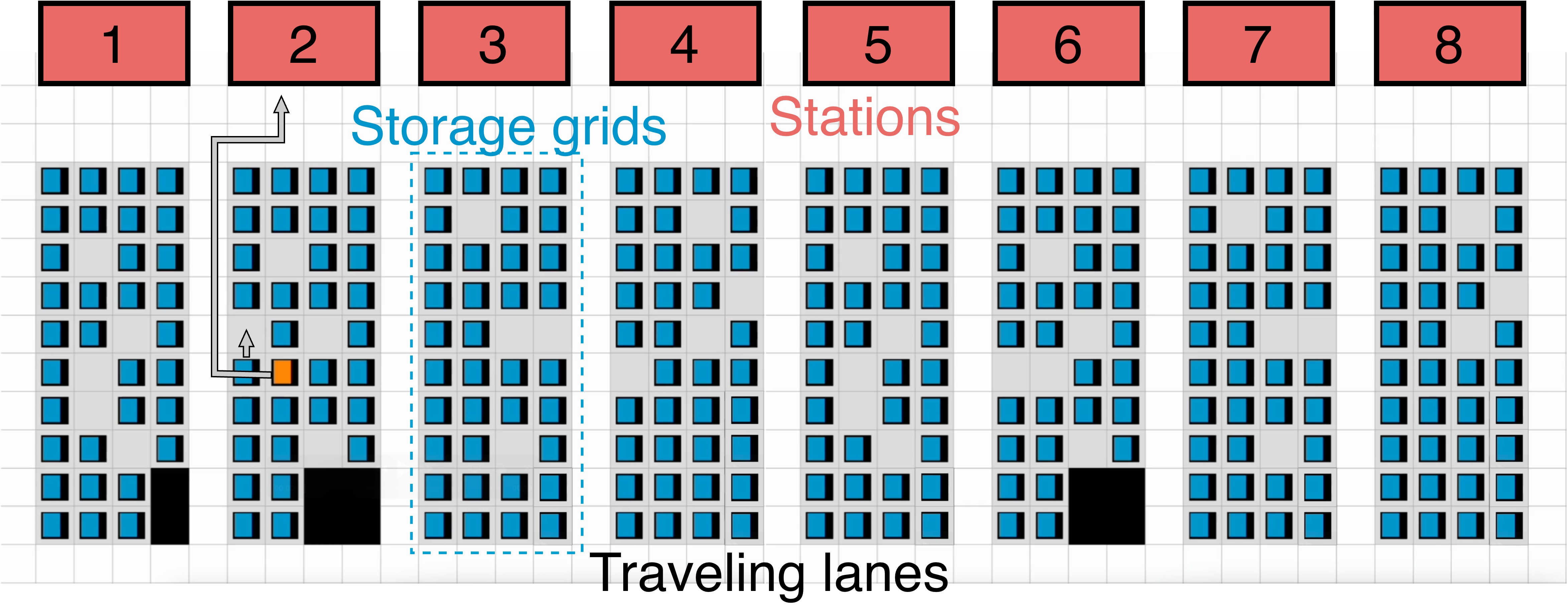}}
    \hfill 
    \caption{Fulfillment center with \blockL{}s in storage \gridL{}s separated by travel lanes. Assigned \blockL{}s marked in orange.}
    \label{fig:fulfillment_center}
\end{figure}

Amazon Robotics fulfillment centers are large warehouses that utilize robotic Automated Storage and Retrieval Systems to store inventory and fulfill customer orders. As illustrated in \figref{} \ref{fig:fulfillment_center}, fulfillment centers store inventory items in shelves (\blockL{}s) arranged into storage \gridL{}s separated by travel lanes. Robot drives can lift and move these \blockL{}s. When a station at the periphery requests an assigned \blockL{} for item picking or stowing, robot drives move that \blockL{}, transport it to the storage \gridL{}'s boundary and then utilize travel lanes to bring it to the station.
The system leverages dense storage \gridL{}s to achieve efficient space utilization and lower fulfillment costs. This creates a challenge: assigned \blockL{}s may be deeply buried and obstructed by other \blockL{}s, requiring complex rearrangement plans.

\begin{figure}[t!]
    \centering
    \subfloat[Definition\label{fig:problem_definition}]{
    \includegraphics[height=0.65\linewidth, trim=0 0 0 0, clip]{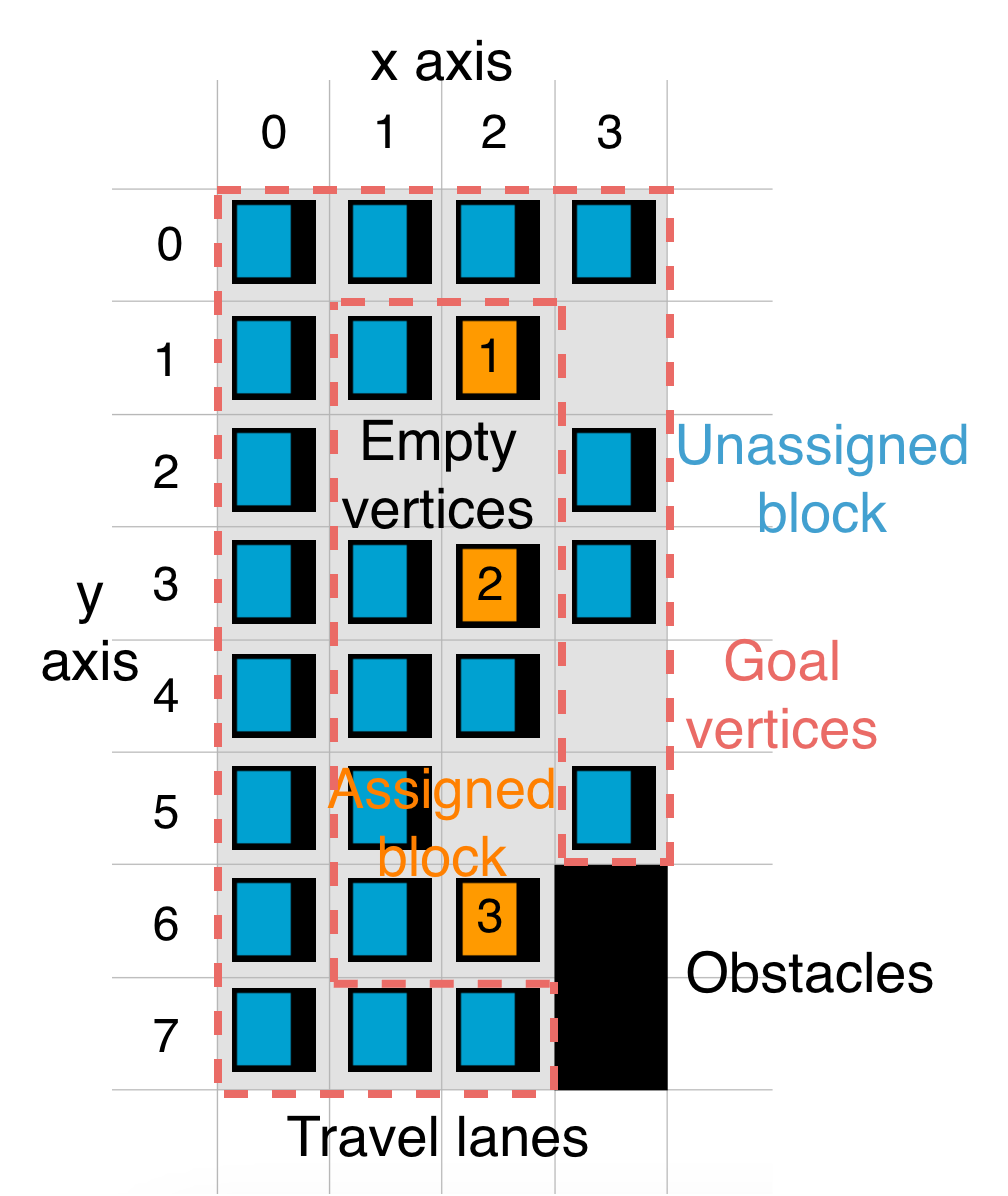}}
    \subfloat[Example plan\label{fig:problem_solution}]{
    \includegraphics[height=0.65\linewidth, trim=0 0 0 0, clip]{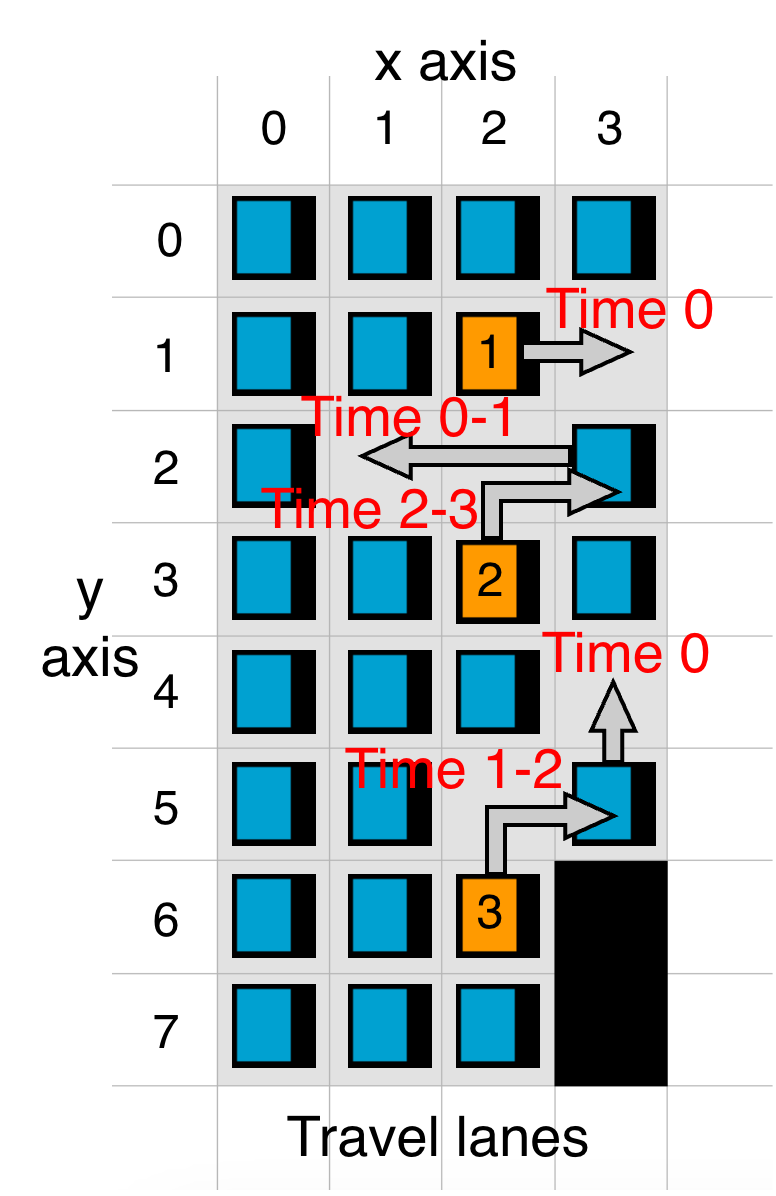}}
    \hfill 
    \caption{\problemupperP{} definition.}
    \label{fig:problem}
\end{figure}

We define the \problemupperP{} (\problemacronymP{}) as follows. Given a \gridL{} layout containing assigned \blockL{}s, unassigned \blockL{}s, empty cells, and obstacles, as illustrated in \figref{} \ref{fig:problem_definition}, the objective is to find the lowest-cost plan to move each assigned \blockL{} to a set of desired goal locations.
Note that the \blockL{}s are also referred to as agents in other problem settings~\cite{mapf}.
These goal locations can be either contiguous or disjoint. When goal locations are at the \gridL{} boundary, assigned \blockL{}s are rearranged for direct travel lane access. Alternatively, when goal locations are within the \gridL{}, assigned \blockL{}s are positioned for potential future access. The plan consists of \blockL{} moving actions with their associated timestamps. Cost metrics can include sum of action costs, makespan, or other user-defined objectives. An example plan for the problem shown in \figref{} \ref{fig:problem_definition} is illustrated in \figref{} \ref{fig:problem_solution}.

The \problemupperP{} in dense warehouses shares fundamental structural similarities with various puzzles and real-world problems, including the sliding tile puzzle \cite{gozon2024computing}, Rush Hour puzzle \cite{cian2022modeling}, block relocation problem \cite{lu2020study}, box world/Sokoban \cite{zawalski2024matters}, parking lot rearrangement \cite{guo2023toward}, Multi-Agent Path Finding (MAPF) \cite{mapf,li2022mapf, Shen2023BeyondPR, lacam}, and MAPF with Unassigned Agents \cite{mapf_ua}.
These problems involve rearrangement in discrete space, aiming to achieve goal configurations under movement constraints.
Due to the combinatorial increase in the number of possible configurations, these problems are generally NP-hard \cite{ratner1986finding, flake2002rush, yu2013structure}.

This paper makes several contributions: it formally defines the \problemupperP{}, establishes similarities and differences between \problemacronymP{} and existing puzzles and research problems, develops five distinct solution algorithms, and evaluates their optimality and computational efficiency. Through this comprehensive analysis, we establish baseline approaches for solving \problemacronymP{}s and identify directions for future research.

% \begin{figure}[t!]
% 	\centering
% 	\includegraphics[width=\linewidth]{example-image-a}
% 	\caption{Fulfillment center with pods in storage \gridL{}s separated by traveling lanes.}
% 	\label{fig:storage_blocks}
% \end{figure}

\section{Problem Description}\label{sec:problem_description}

A \problemacronymP{} is defined on a graph \(G = (V, E)\) with the set of vertices \(V\) and edges \(E = \{(u, v) \ | \ u, v \in V\}\) connecting adjacent vertices. Let \(\Ical = \{1, \cdots, n_{\Ical}\}\) denote the set of assigned \blockL{}s (marked in orange in \figref{} \ref{fig:problem}) and each assigned \blockL{} \(i \in \Ical\) starts at an initial vertex \(s_i \in V\) and must be moved to a set of desired goal vertices \(V_i \subseteq V\). Additionally, \(\Jcal = \{n_{\Ical} + 1, \cdots, n_{\Ical} + n_{\Jcal}\}\) denotes unassigned \blockL{}s without desired goals.
Actions \(A = \{\move{}, \wait{}, \complete{}\}\) comprise a set of operations that modifies the time or location of the \blockL{}s. A path \(p_i\) for \blockL{} \(i \in \Ical \cup \Jcal\) consists of an action sequence \((a_i^1, a_i^2, \cdots, a_i^{|p_i|}), a_i^k \in A\). Assigned \blockL{} paths must move assigned \blockL{}s to goal vertices to be completed, while unassigned \blockL{} paths facilitate assigned \blockL{} movements.
A solution \(P = \{p_i \ | \ i \in \Ical \cup \Jcal\}\) to a \problemacronymP{} is a set of conflict-free paths, that transition all the assigned \blockL{}s to their goal vertices. 
The objective is to find a feasible solution minimizing either sum of action costs, makespan, or other user-defined cost metrics, as detailed in Sec. \ref{sec:problem-objectives}.

\subsection{Actions}\label{sec:problem-actions}

The action set \(A\) contains three types of actions that can be applied to the \blockL{}s, each takes one time step to execute and is associated with a corresponding cost. The action \move{\((i, t, u, v)\)} moves \blockL{} \(i \in \Ical \cup \Jcal\) between two adjacent vertices \((u, v) \in E\) at time step \(t\) if \(u\) is a \blockL{} and \(v\) is an empty vertex. \wait{\((i, t)\)} forces a \blockL{} to wait in place at time \(t\). The \complete{\((i, t, v)\)} action can only be applied when an assigned \blockL{} \(i \in \Ical\) is at one of its goal vertices \(v \in V_i\). The complete action removes the \blockL{} from the assigned \blockL{} set: \(\Ical \leftarrow \Ical \backslash i\).

\begin{figure}[t!]
    \centering
    \hfill
    \subfloat[Vertex]{\includegraphics[width=0.16\linewidth]{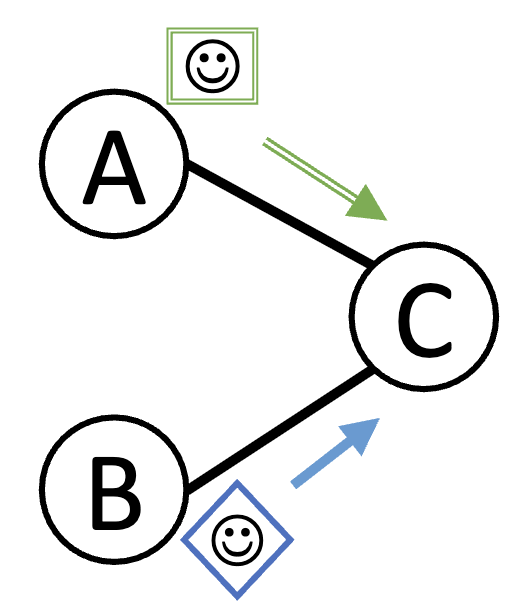}\label{fig:infeasible-movement-vertex}}
    \hfill
    \subfloat[Edge]{\includegraphics[width=0.18\linewidth]{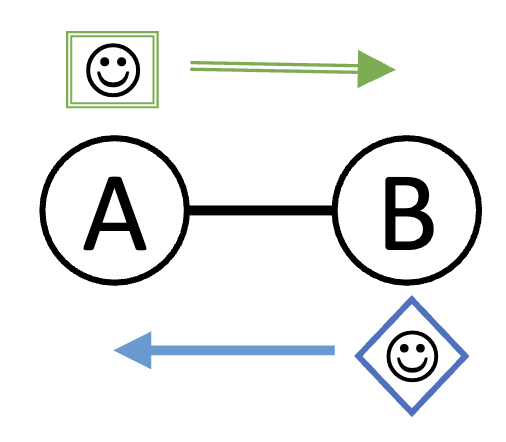}\label{fig:infeasible-movement-edge}}
    \hfill
    \subfloat[Following]{\includegraphics[width=0.27\linewidth]{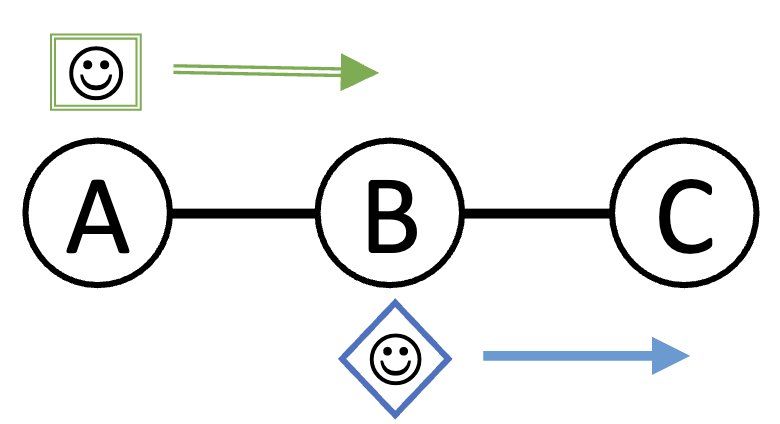}\label{fig:infeasible-movement-following}}
    \hfill
    \caption{Three types of conflicts~\cite{mapf}.}
    \label{fig:infeasible-movement}
\end{figure}

It is important to note that, according to the above definition, the destination of a \move{} action must be an empty vertex. We prohibit platooning, where one \blockL{} closely follows another, to better align with warehouse operational constraints. Specifically, the movement must avoid three conflicts~\cite{mapf}:
\begin{itemize}
    \item \emph{Vertex Conflict} (\figref{} \ref{fig:infeasible-movement-vertex}), where two \blockL{}s use the same vertex at the same timestep,
    \item \emph{Edge Conflict} (\figref{} \ref{fig:infeasible-movement-edge}), where two \blockL{}s use the same edge at the same timestep,  
    \item \emph{Following Conflict} (\figref{} \ref{fig:infeasible-movement-following}), where one \blockL{} occupies a vertex that was occupied by another \blockL{} in the previous time step.
\end{itemize}

\subsection{Terminal State}\label{sec:problem-termination}

The objective is to empty the assigned \blockL{} set through completion actions.
Three distinct completion actions are possible for assigned \blockL{}s. In the first case, an assigned \blockL{} reaching the goal (e.g., the boundary) is considered rearranged and removed, converting its location to an empty vertex. In the second case, an assigned \blockL{} becomes unassigned upon reaching its goal location. In the third case, an assigned \blockL{} becomes an obstacle at its goal location. Each completion action corresponds to a specific practical scenario. This paper focuses on developing algorithms and conducting experiments using the third option.

\subsection{Objectives and Metrics}\label{sec:problem-objectives}

Multiple metrics can be used to evaluate the quality of the \problemacronymP{} solutions, including the number of moves of assigned \blockL{}s, the number of moves of unassigned \blockL{}s, sum of action costs, and the makespan of the whole plan. In \problemupper{}, we care about all of these metrics.

Let \(c(a)\) denote the cost of an action \(a \in A\).
\(c(a)\) returns the cost according to the types of the \blockL{}s and the actions. An example action cost matrix is illustrated in \tabref{} \ref{tab:action-cost}.

\begin{table}[htbp]
  \small
  \centering
    \begin{tabular}{c|ccc}
    \toprule
    \blockU{} type & \move{}  & \wait{}  & \complete{} \\
    \midrule
    Assigned & 2     & 1     & 2 \\
    Unassigned & 2     & 0     & N/A \\
    \bottomrule
    \end{tabular}%
  \caption{Example action costs}
  \label{tab:action-cost}%
\end{table}%

The cost of a path \(p_i = (a_i^1, \cdots, a_i^{|p_i|})\) is defined as:
\begin{align}
    c(p_i) = &\sum_{k = 1}^{|p_i|} c(a_i^k) \label{eqn:path_cost} 
\end{align}
We define the objectives: composite cost of a solution  \(P = \{p_i \ | \ i \in \Ical \cup \Jcal\}\) and the makespan of the solution, according to \eqref{eqn:solution_cost}-\eqref{eqn:solution_makespan}, respectively.
\begin{align}
    \text{Composite cost: } c(P) = &\sum_{i \in \Ical \cup \Jcal} c(p_i) \label{eqn:solution_cost}\\
    \text{Makespan: } c_m(P) = &\max_{i \in \Ical \cup \Jcal} c(p_i)
    \label{eqn:solution_makespan}
\end{align}

\subsection{Variants of the Problem}

The Problem can be extended in several ways to accommodate different operational requirements:

\begin{itemize}
\item \textit{Constraint on simultaneous actions:} Limiting the number of \blockL{}s that can move concurrently to reflect real-world constraints.

\item \textit{Shared or distinct goal sets for \blockL{}s:} \blockU{}s might have individual specific goals or share a common set of acceptable end positions.

\item \textit{Priority assignments:} Additional priority for certain \blockL{}s, ensuring critical blocks are rearranged first.

\end{itemize}

\section{Related Work}\label{sec:related_work}

As mentioned in Sec. \ref{sec:introduction}, \problemacronymP{} is closely related to various puzzles and research problems. 
Solutions to these problems have employed diverse approaches including symbolic planning \cite{davesa2024towards}, search and conflict resolution \cite{li2022mapf, sharon2015conflict, ma2019searching}, mixed-integer programming \cite{guo2023toward, lu2020study}, boolean satisfiability \cite{cian2022modeling}, reinforcement learning \cite{shoham2021solving, damani2021primal, wang2025lns2+}, and supervised learning formulations \cite{li2020graph, jiang2025deploying} to find low-cost plans.
Given their demonstrated optimality and efficiency, we leverage symbolic planning and MAPF algorithms to develop the planning methods in this paper.

\subsection{Symbolic Planning and PDDL}

Symbolic planning is a model-based approach in artificial intelligence that solves decision-making problems by explicitly representing states, actions, and their transitions using symbolic logic. Planning languages like Planning Domain Definition Language (PDDL) enable the formulation of complex planning problems as structured domains, encompassing crucial elements such as initial conditions, feasible actions, and desired goal states, allowing planners to search for action sequences that achieve specified goals.
PDDL has been successfully applied to various domains: modeling Rush Hour and sliding puzzles \cite{davesa2024towards}, formulating Rubik's cube solutions \cite{muppasani2024solving}, solving logistics problems for transport routing \cite{helmert2009concise}, and finding robot paths in grid worlds \cite{estivill2013path}. Symbolic representations have also proven effective in task and motion planning for robotic manipulation \cite{garrett2021integrated, silver2021learning}. Recent work has combined the symbolic representation with deep learning and large language models to create flexible long-horizon plans \cite{srivastava2022behavior, agia2023stap, lin2023text2motion}. Building on this foundation, we apply symbolic planning in configuration space and PDDL descriptions to develop planning algorithms for \problemacronymP{}s.

\subsection{Multi-Agent Path Finding}

Multi-Agent Path Finding~\cite{mapf} is a problem of navigating a team of agents from their start locations to goal locations without collisions.
The \problemacronymP{} can be modeled as a complex MAPF variant that combines elements from the classic MAPF problem~\cite{mapf}, where agents (\blockL{}s) must be moved to dedicated goal locations, the anonymous MAPF problem~\cite{mapf, amapf_yu}, where goal vertices are unlabeled and are interchangeable for assigned \blockL{}s,
and the Graph Motion Planning Problem~\cite{graph_motion}, where unassigned \blockL{}s exists and must be moved to clear paths for assigned \blockL{}s but is limited to one motion per step.
More recently, \citet{mapf_ua} introduced the Multi-Agent Path Finding with Unassigned Agents (MAPFUA), which formally defines this setting where assigned agents (our assigned blocks) have goals and others (our unassigned blocks) can move to clear the way.
Our BRaP can be seen as a specific, challenging variant of MAPFUA, characterized by extremely dense environments, strict movement constraints, and task assignments.
For a broader discussion of the MAPFUA problem and its potential variants, we direct readers to the MAPFUA paper~\cite{mapf_ua}.

Different from the anonymous MAPF problem, assigned \blockL{}s may be assigned distinct goal vertex sets thus goal vertices are not fully interchangeable.
This aspect is related to the Target Assignment and Path Finding (TAPF) problem~\cite{tapf}, where agents in a team can be assigned to any of one of the team's dedicated goal vertices. However, the TAPF problem does not include unassigned agents or blocks.

The \problemacronymP{} is made further challenging by the high-density environment, where unassigned \blockL{}s often occupy the vast majority of the workspace, and by its specific operational constraints prohibiting following.
Although ``following conflicts" are defined in the literature, most solvers ignore them and it is non-trivial to incorporate them into certain MAPF algorithms.
These combined factors mean that even though powerful, state-of-the-art solvers like \lacam{}~\cite{lacam,lacam2,lacam3} are effective for high-density MAPF problems, they are not directly applicable to \problemacronymP{}s.
Therefore, building upon these advanced frameworks, we develop a MAPF-based solution tailored to the specific complexities of the \problemacronymP{}.

\section{Applicable Methodologies}\label{sec:methods}
\subsection{Configuration Space Search}\label{sec:method-configuration-search}

The \problemacronymP{} can be formulated as a graph search in configuration space. Starting from an initial \gridL{} configuration, available actions transform the state until reaching a goal configuration that satisfies desired criteria. Actions at each time step represent edges that transition between neighboring configuration states. Each state comprises time, assigned \blockL{} set (vertices containing assigned \blockL{}s), empty vertex set, and completed \blockL{} set (vertices containing completed assigned \blockL{}s).

\begin{figure}[t!]
    \centering
    \includegraphics[width=\linewidth]{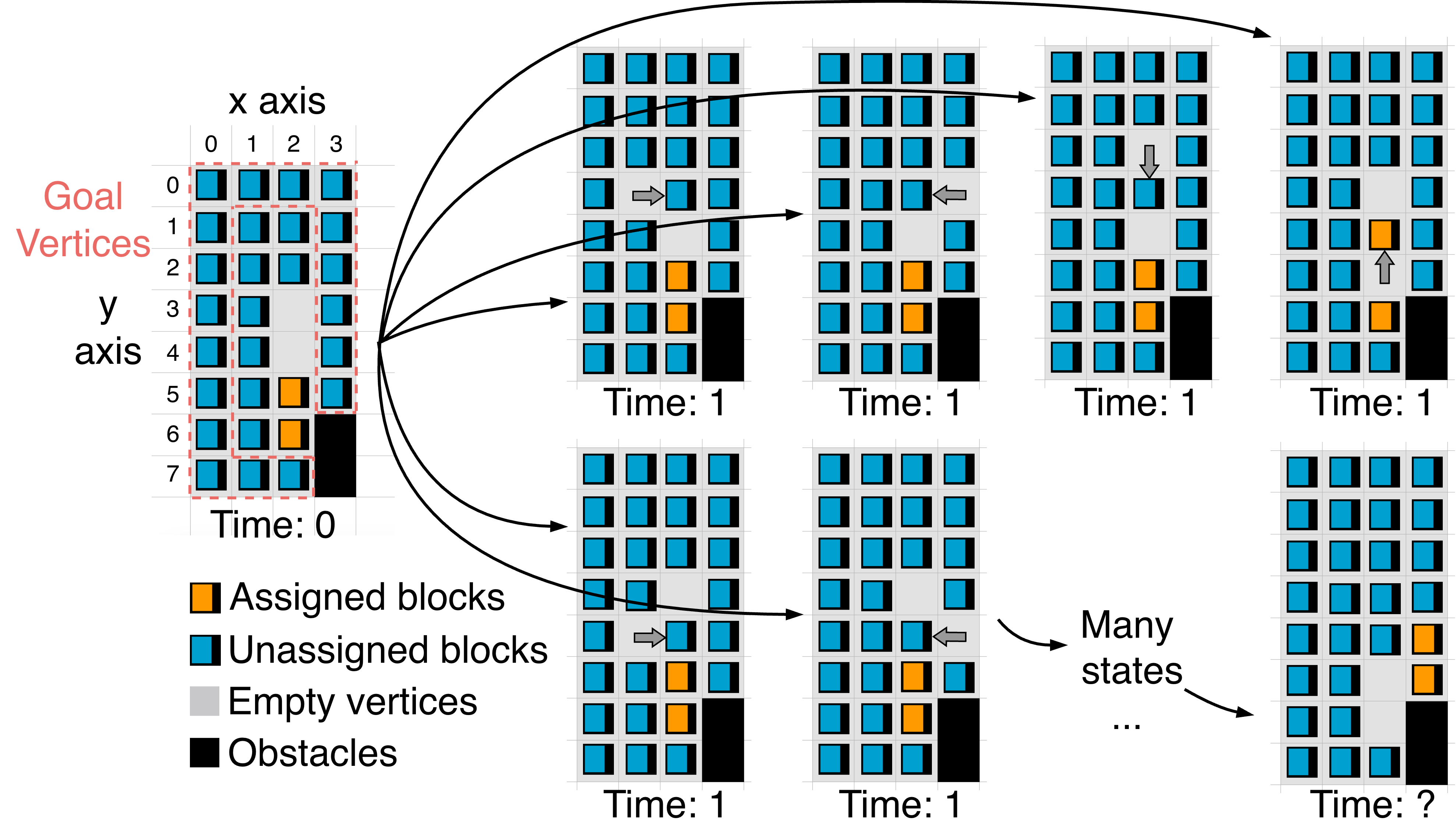}
    \caption{A graph of configuration states and actions.}
    \label{fig:configuration-states}
\end{figure}

An example graph of configuration states and actions is illustrated in \figref{} \ref{fig:configuration-states}. According to the above definition, the root state in the graph is: time: 0, assigned \blockL{}s: \{(2,5), (2,6)\}, empty vertices: \{(2,3), (2,4)\}, completed \blockL{}s: \(\emptyset\). The assigned \blockL{} set in the goal state should be empty.

The configuration space can be modified using the three actions described in Sec. \ref{sec:problem-actions}. To ensure computational tractability and reduce the branching factor, we restrict the system to one action per time step. This restriction renders the \wait{} action redundant and thus disallowed. Methods supporting multiple actions at each time step will be discussed in subsequent sections.

A configuration graph can be constructed from the defined states and actions, as shown in \figref{} \ref{fig:configuration-states}.
With the cost of the actions defined, graph search techniques like Dijkstra and A* algorithms can be applied to find the minimum cost path from the initial state to the goal state (where there is no longer any assigned \blockL{}s).

A* with an admissible heuristic function is guaranteed to be complete (always find a solution if there is) and optimal.
We formulate an admissible heuristic by assuming assigned \blockL{}s move along least-blocking vertices, with each blocking \blockL{} requiring only one \move{} action for removal. The detailed heuristic formulation is presented in Appendix.
% ~\ref{appendix:heuristic}.

\subsection{PDDL-based Configuration Space Search}

The configuration space search can also be described using PDDL. We define five predicates: \texttt{(emp ?u)}, \texttt{(asb ?u)}, \texttt{(blk ?u)}, \texttt{(cmp ?u)}, and \texttt{(goal ?u)}, indicating whether a vertex \(u \in V\) contains an empty space, assigned \blockL{}, unassigned \blockL{}, completed \blockL{}, or potential goal, respectively. The predicate \texttt{(adjacent ?u ?v)} denotes that vertices \(u\) and \(v\) are adjacent to each other, i.e., \((u,v) \in E\).
Three actions are defined in the PDDL domain: \texttt{(move\_blk ?u ?v)} and \texttt{(move\_asb ?u ?v)} for moving unassigned and assigned \blockL{}s, respectively, and \texttt{(complete ?u)} for completion.
% The full PDDL domain is detailed in Appendix~\ref{appendix:pddl}.

In the problem file, vertices serve as objects, denoted as \texttt{node-0-0}, \texttt{node-0-1}, etc., excluding vertices containing obstacles. The initial state specifies each node's type as either \texttt{(asb node-x-y)}, \texttt{(blk node-x-y)}, or \texttt{(emp node-x-y)}. All vertices except assigned \blockL{}s are marked with \texttt{(cmp node-x-y)}, and valid terminal locations are marked with \texttt{(goal node-x-y)}. The goal state requires all vertices to be marked as \texttt{(cmp node-x-y)}, with total action cost as the optimization metric.

While this formulation allows simultaneous moves, the \texttt{fast-downward} solver used in Sec. \ref{sec:results} generates totally-ordered plans with single actions per timestep. Formulations that support multi-agent temporal planning would introduce prohibitively high branching factors for PDDL solvers, motivating the restriction to single-action plans.

\subsection{Priority-based Configuration Space Search}\label{sec:method-priority-search}

In contrast to the single action per time step approach of Sec. \ref{sec:method-configuration-search}, we now develop a formulation allowing one action per assigned \blockL{} per time step, enabling parallel execution and reducing rearrangement makespan.

Algorithm \ref{alg:priority-based-search} begins by assigning priorities based on \blockL{} proximity to goals, as our experiments showed this produces better solutions. The priority for each assigned \blockL{} \(i \in \Ical\) is determined using the heuristic \(h(s, i)\) described in Appendix, where smaller values indicate higher priorities.

The search iteratively creates rearrangement plans for individual assigned \blockL{}s while respecting plans of higher-priority \blockL{}s as constraints. For a prior plan \(P\), the \textit{computeConstraint}(\(P\)) function generates constraints ensuring the preconditions of higher-priority actions are satisfied. For example, if \move{\((i, t, u, v)\)} appears in \(P\), a constraint in \(\xi\) ensures \blockL{} \(i\) occupies \(u\) at time \(t\) while \(v\) remains empty, and no other actions can modify these vertices at time \(t\).
The \textit{computeConstrainedPlan}() function generates plans by pruning states violating \(\xi\) and treating prior plans as external forces, ensuring parallel executability of the resulting plans.

\begin{algorithm}[t]
% \SetAlgoLined
% \SetAlgoNoLine
\small

\textbf{Input:} \gridU{} \(G(V,E)\), assigned \blockL{} set \(\Ical\), goal vertices \(V_i (i \in \Ical)\)

\(\Ical_{\text{sorted}} \leftarrow\) \textit{computePriority}(\(\Ical, V_i, V, E\))

\(P \leftarrow \emptyset\) \CommentSty{//The initial plan}

\For{\(i \in \Ical_{\text{sorted}} \)}{
    \(\xi \leftarrow\) \textit{computeConstraint}(\(P\))

    \(p_i \leftarrow\) \textit{computeConstrainedPlan}(\(i, V_i, V, E, P, \xi\))

    \(P \leftarrow P \cup p_i \)
   
}

\Return \(P\)

\caption{Priority-based search}
\label{alg:priority-based-search}
\end{algorithm}

\subsection{Heuristic Approach: Moving Along the Least Blocking Path}

Given the exponentially growing configuration search space with respect to the block number, algorithms in the previous three sections do not scale efficiently to large \gridL{}s with numerous assigned \blockL{}s. In this section, we propose a pure heuristic-based approach that moves each assigned \blockL{} along the least-blocking vertex path.

\begin{figure}[t!]
    \centering
    \includegraphics[width=0.7\linewidth]{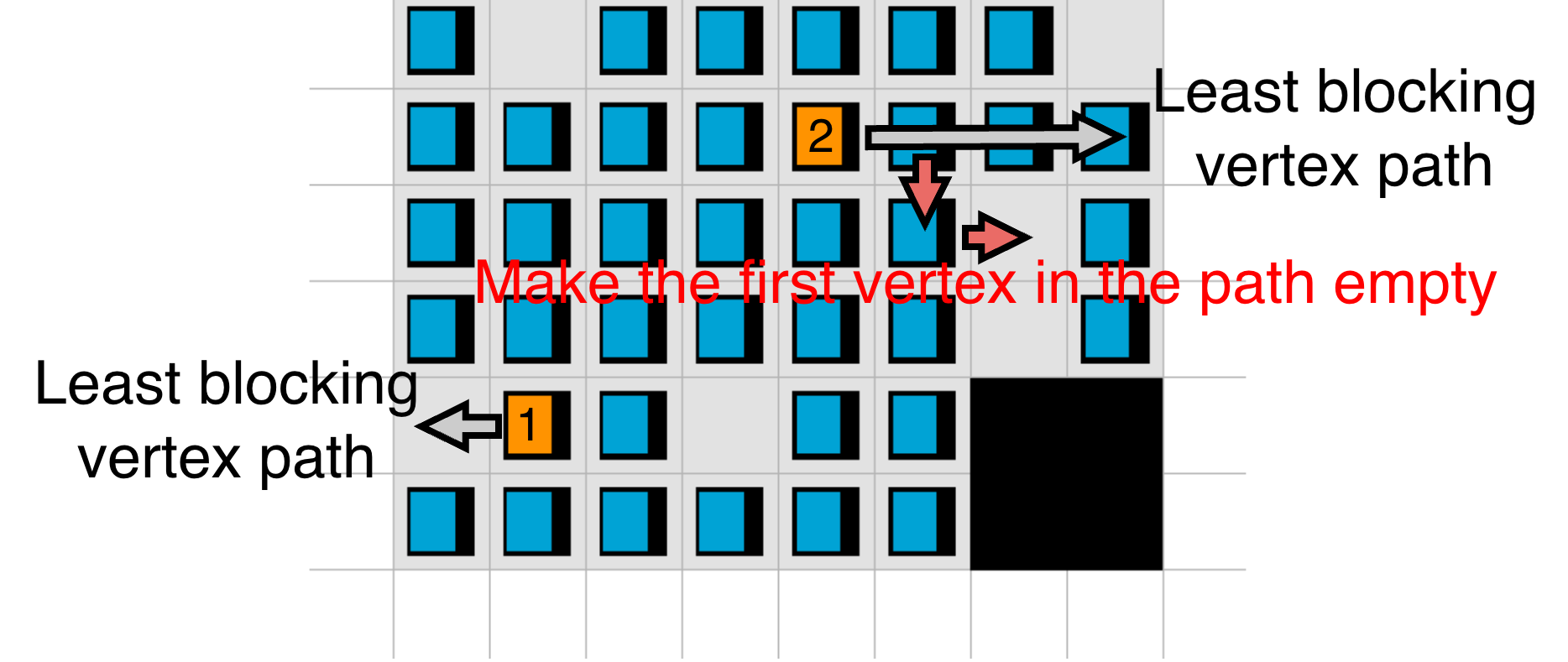}
    \caption{Move assigned \blockL{}s along the least blocking path.}
    \label{fig:problem-heuristic}
\end{figure}

Algorithm \ref{alg:heuristic-algorithm} outlines the heuristic approach. The process begins by allocating priorities to assigned \blockL{}s using the method described in Sec. \ref{sec:method-priority-search}. As illustrated in \figref{} \ref{fig:problem-heuristic}, for each assigned \blockL{}, the planner first creates a least-blocking path, \(U_i = \{u_i^1, \cdots, u_i^{|U_i|}\}\), to the goal vertices. Then, for each vertex along this path, \(u_i^l \in U_i\), the planner create an empty vertex at \(u_i^l\) by iteratively moving blocks to the closest empty vertex and subsequently moves the assigned \blockL{} to the emptied vertex \(u_i^l\), thus generating a \blockL{} movement plan. A least-blocking path is defined as the lowest-cost path where assigned \blockL{}s, unassigned \blockL{}s, and empty vertices are assigned different traversal costs.

During planning, the plan for the current assigned \blockL{} \(i\) is created based on the map status where all previous assigned \blockL{} plans have been executed. However, executing the plan for assigned \blockL{} \(i\) does not require waiting for complete execution of all previous plans. Instead, it only requires waiting for the completion of plans affecting vertices that will be used by plan \(p_i\). The function \(t = \psi(u)\) takes a vertex \(u \in V\) and returns this waiting time. By ensuring actions in \(p_i\) occur no earlier than times specified by \(\psi\), the planner enables parallel movements of multiple assigned \blockL{}.

\begin{algorithm}[t]
% \SetAlgoLined
% \SetAlgoNoLine
\small

\textbf{Input:} \gridU{} \(G(V,E)\), assigned \blockL{} set \(\Ical\), goal vertices \(V_i (i \in \Ical)\)

\(\Ical_{\text{sorted}} \leftarrow\) \textit{computePriority}(\(\Ical, V_i, V, E\))

\(P \leftarrow \emptyset\) \CommentSty{//The initial plan}

\For{\(i \in \Ical_{\text{sorted}} \)}{
    \(\psi \leftarrow\) \textit{computeVertexBlockingTime}(\(P\))

    \(U_i \leftarrow\) \textit{leastBlockingPath}(\(i, V_i, V, E, P\))

    \(p_i \leftarrow \emptyset\)
    
    \For{\(u \in U_i \)}{
        \(p_i^u \leftarrow\) \textit{createEmptyVertexAt}(\(u, V, E, P, \psi\))
    
        \(p_i \leftarrow p_i \cup p_i^u \cup \move{(i, t, u_{\text{previous}}, u)} \)
    }
    
    \(P \leftarrow P \cup p_i \cup \complete{(i, t, u)}\)
   
}

\Return \(P\)

\caption{Heuristic approach}
\label{alg:heuristic-algorithm}
\end{algorithm}

\subsection{LaCAM-based Multi-Agent Path Finding}
By discretising time into timesteps and allowing multiple \blockL{}s to move to empty vertices simultaneously in each timestep, the problem is similar to a Multi-agent Path Finding problem. 
With the recent advances in solving large scale and high density MAPF problems, we introduce Block Rearrangement \lacam{}, a \emph{Lazy Constraints Addition Search} (\lacam{}) \cite{lacam,lacam2,lacam3} based algorithm to solve the \problemupperP{} in this section.

\subsubsection{Lazy Constraints Addition Search}
\lacam{} is a heuristic search algorithm that aims to rapidly find an initial solution and then continuously search for lower cost ones. 
The algorithm has three major components: (1) the high-level search tree, (2) the configuration generator, and (3) the low-level constraints.
Similar to the Configuration Space Search presented in Sec. \ref{sec:method-configuration-search} and \figref{} \ref{fig:configuration-states}, the high-level search component of \lacam{} searches over the configuration space for solutions. 
A high-level search node $n$ in \lacam{} not only stores a configuration state $n.\pi$, but also low-level constraints $n.\xi$ which store spatial constraints that force the constrained \blockL{}s to perform the constrained actions.
To rapidly find an initial solution, \lacam{} searches in a way that combines \emph{Depth-first Search} (DFS) and \emph{Greedy Best-first Search} (GBFS).
 
However, unlike DFS that blindly expands from one node to the next or GBFS which applies a heuristic function to evaluate all possible successor states and expands the most promising one,
\lacam{} utilizes a configuration generator $f(n)$ to expand a given node $n$ and generate a single successor node $n'$ for exploration.
$f(n)$ itself is a lightweight one step planner (e.g. PIBT~\cite{pibt}) which suggests the most promising next state without enumerating all possible successor states.

If the new configuration, $n'.\pi$, has not yet been explored, the algorithm advances to $n'$ and continues its deep dive in a DFS-style exploration, until a goal node is found.
Conversely, if $n'.\pi$ has already been explored, the algorithm backtracks to the parent node $n$ and grows its constraints, $n.\xi$. 
This addition forces the configuration generator, $f(n)$, to generate an alternative successor configuration state. 
$n.\xi$ grows in a brute-force manner to gradually take over the control of \blockL{}s from the configuration generator, and eventually force the exploration of all possible successor nodes of $n$ if $f(n)$ keeps returning explored states.

% ============================================
% ============ Begin of algorithm ============
% ============================================
\begin{algorithm}[t]
\small

\textbf{Input:} \gridU{} $G(V,E)$, assigned \blockL{} set $\mathcal{I}$, unassigned set $\Jcal$, goal vertices $V_i$ $(i \in \mathcal{I})$, priorities $\omega$, current states $\pi$, next states $\pi'$, current temp goals $g$, 

% $\pi_i' \leftarrow $ \llIf{$c_i \neq null$}{$c_i$} \llElse{$\perp$}: for each $i \in \Ical\cup\Jcal$\;
\textit{UpdatePriorities}($\mathcal{I}, V_i, \omega, \pi$)\; \label{alg:DSR-PIBT:update-priorities}

$g' \leftarrow \emptyset$; \CommentSty{//Next temp goals}\\
% \For{$i \in \Ical$}{
%     \llIf{$\pi_i = g_i$}{$g_i \leftarrow \text{null}$}\label{alg:DSR-PIBT:goal-release}; \CommentSty{//Release temp goal}\\
%     $g_i' \leftarrow null$; \CommentSty{//Next temp goal}\\
% } 

\For{each \blockL{} $k \in \Ical$ in descending priority order}{ \label{alg:DSR-PIBT:priority-order}
    \lIf{$\pi_k' = \perp$}{
        $\textit{BR\_PIBT}(\text{null}, k)$ 
    }
    \label{alg:DSR-PIBT:priority-order-end}
}

\Return $\pi'$, $g'$

\textbf{\textit{BR\_PIBT}}(parent, \blockL{} $i$): \label{alg:DSR-PIBT:start}
\Begin{
    % $\text{Proccessed}(\pi_i)$; 
    $\pi_i' \leftarrow \pi_i$; \CommentSty{//Default action wait}\\
    % \\
    
    \lIf{$i \in \Ical$}{
         $g_i' \leftarrow$ \llIf{$g_i$ is null or occupied in $g'$}{
           $\text{closest\ goal\ }v \in V_i$, $v \notin g'$ \label{alg:DSR-PIBT:goal-take-away}
        }
        \llElse{
            $g_i$
        }
    }
    % $\text{require\_assignment} \leftarrow g_i = \text{null}$\;

    \lIf{no unused empty vertices left}{
        \Return true
    }
    $C \leftarrow \text{neighbors}(\pi_i) \cup \{\pi_i\}$;  \CommentSty{//Candidate locations} \label{alg:DSR-PIBT:assign-metrics}\\
 
    \lFor{each $u \in C$}{
        \textit{AssignPreferences}($u, i, g'_i$)\label{alg:DSR-PIBT:sort-metrics}
    }
    
    Sort $C$ by preference in ascending order\;
    \For{each candidate $u \in C$}{
        \lIf{$u$ is occupied in $\pi'$}{
            \textbf{continue}
        }
        \lIf{$parent \neq null$ and $u = \pi_i$}{\textbf{continue}\label{alg:DSR-PIBT:skip-wait}}

        $j \leftarrow$ the \blockL{} occupies $u$ in $\pi$\;        
        
        \If{$j \neq null$ and $j \neq i$}{
            \lIf{ $\pi'_j \neq \perp$ or not $\textit{BR\_PIBT}(i, j)$}{\textbf{continue}}
        }
        \llIf{$j = null$}{
            $\pi_i' \leftarrow u$;
        } \CommentSty{//Move if empty}\\

        \Return true\;
    }
    
    \Return false\; \label{alg:DSR-PIBT:end}
}

\textbf{\textit{UpdatePriorities}}($\mathcal{I}, V_i, \omega, \pi$): \label{alg:DSR-PIBT:update-priorities-definitions}
\Begin{
    \For{each $i \in \mathcal{I}$}{
    $\omega_i \leftarrow$ \llIf{ $\pi_i \in V_i$ or $\omega_i = null$}{ $rand(0,1)$}
    \llElse{$\omega_i + 1.0$}
    }
}

\textbf{\textit{AssignPreferences}}($u, i, g'_i$): \label{alg:DSR-PIBT:assign-preference-definitions}
\Begin{
    $u.preference \leftarrow$ \llIf{$i \in \mathcal{I}$}{$(Distance(u, g_i'), DistanceToEmpty(u))$} \llElse{$(DistanceToEmpty(u), random())$}
}

\caption{\brpibt{}}
\label{alg:DSR-PIBT}
\end{algorithm}

% ============================================
% ============ End of algorithm ==============
% ============================================

\subsubsection{\brpibt{} Configuration Generator}

With the existence of following conflicts, PIBT~\cite{pibt} is no longer applicable as a configuration generator for 
\lacam{} in \problemacronymP{}. 
This is because PIBT recursively decides the actions for chains of movable \blockL{}s, while, due to following conflicts, only the \blockL{}s next to empty vertices are movable.
The additional complexity is the \problemacronymP{} has multiple goal vertices for each assigned \blockL{}, and the configuration generator needs to decide which goal vertex to move towards.
In this section, we introduce \brpibt{} as the configuration generator for Block Rearrangement \lacam{} (\brlacam{}).

The core intuition of the \brpibt{} algorithm is that every \blockL{} takes turns to call for empty vertices in a priority-based manner, ensuring that each \blockL{} is processed exactly once per step to maintain algorithmic efficiency. 
This approach ensures that all \blockL{}s have the opportunity to make progress toward their goals by dynamically updating their priorities. \algof~\ref{alg:DSR-PIBT} describes the process of \brpibt{}.

The \brpibt{} algorithm operates as a single-step planner that generates the next configuration state by iteratively assigning actions and temporary goals to \blockL{}s based on dynamically managed priorities. The algorithm maintains three key components: \textit{1) Priority Management, 2) Recursive Planning and 3) Temporary Goal Allocation}.

\texttt{1) Priority Management:} Similar to PIBT, each \blockL{} $i$ maintains a priority value $\omega_i$ that determines the order in which \blockL{}s are processed. Priorities are dynamically updated based on goal achievement and movement status. When a \blockL{} $i$ reaches any vertex in $V_i$, its priority is reset to a floating number in $(0,1)$. When the \blockL{} $i$ is displaced from its goal vertices $V_i$, its priority is incremented by $1$. This priority update process happens at the beginning of each configuration generation phase (\algof~\ref{alg:DSR-PIBT} line~\ref{alg:DSR-PIBT:update-priorities} and \ref{alg:DSR-PIBT:update-priorities-definitions}).

\texttt{2) Recursive Planning:}
The algorithm processes assigned \blockL{}s according to their priority (\algof~\ref{alg:DSR-PIBT} lines~\ref{alg:DSR-PIBT:priority-order} -~\ref{alg:DSR-PIBT:priority-order-end}). For each \blockL{}, it attempts to move to the most preferred adjacent location. If the desired spot is occupied, the algorithm makes a recursive call on the blocking \blockL{}, finding a chain of displacement requests that reaches an empty vertex (\algof~\ref{alg:DSR-PIBT}, lines~\ref{alg:DSR-PIBT:start} -~\ref{alg:DSR-PIBT:end}).
During this process, the algorithm returns the default action \wait{} if no empty vertices are left or the chain of displacement is not possible. 
It then identifies available neighboring vertices and sorts them based on action preference metrics (\algof~\ref{alg:DSR-PIBT} lines~\ref{alg:DSR-PIBT:assign-metrics} -~\ref{alg:DSR-PIBT:sort-metrics}). 
The \wait{} action is only considered as a preferred option if the \blockL{} is not being actively pushed by another \blockL{} to find the chain (\algof~\ref{alg:DSR-PIBT} line ~\ref{alg:DSR-PIBT:skip-wait}). 
The process ensures that a \blockL{} remains stationary unless a move is strategically advantageous for itself or required to accommodate another \blockL{}'s movement.

\texttt{3) Temporary Goal Allocation:} For assigned \blockL{}s, \brpibt{} allocates temporary goals to guide them. 
When a \blockL{} decides its action, a temporary goal is selected from the closest unallocated goal locations if the \blockL{} does not have one or if its goal was taken away by a higher priority \blockL{} (\algof~\ref{alg:DSR-PIBT} line~\ref{alg:DSR-PIBT:goal-take-away}).

The action preference differentiates between assigned \blockL{}s and unassigned \blockL{}s (\algof~\ref{alg:DSR-PIBT} line~\ref{alg:DSR-PIBT:assign-preference-definitions}). Assigned \blockL{}s prioritize moves that minimize the distance to their goals, while also considering proximity to empty vertices for tie breaking. Unassigned \blockL{}s focus primarily on maintaining access to empty vertices to facilitate overall system mobility. 
The algorithm's efficiency stems from its guarantee that each \blockL{} is processed at most once per timestep. When a \blockL{} is called recursively to facilitate movements, it is immediately decided an action, thus is marked as processed.

\subsubsection{Integration with LaCAM}

The \brpibt{} configuration generator integrates seamlessly with the \lacam{} framework.
For each high-level node expansion, \lacam{} uses \brpibt{} to generate a single successor configuration state.
If explored state are generated and low-level constraints are imposed, \lacam{} forces \brpibt{} to satisfy the low-level constraints by first setting the next states in $\pi'$ as the constrained locations of constrained \blockL{}s, then calling \brpibt{} for decisions on unconstrained \blockL{}s. 
The combination of \lacam{} and \brpibt{}, called \brlacam{}, enables the rapid exploration of the configuration space while still maintaining \lacam{}'s completeness guarantee.

\textbf{Claim 1}: \brlacam{} is complete, which returns a solution for solvable \problemupperP{}s.
\begin{proof}
Following \citep{lacam}, the search space is finite and the number of configuration states is $|V|^{|\Ical \cup \Jcal|}$.
Since the low-level constraints enumerate all possible combinations of movements for all \blockL{}s, it generates all possible configuration states that connects to a high-level node's configuration state. 
Consequently, all reachable configuration states from the start configuration state are explored.
\end{proof}

\subsubsection{Anytime Improvement}

The other advantage of \lacam{} is its anytime search ability.
After an initial solution is found, it uses this initial solution cost as an upper bound and continues to explore the search space for a lower cost solution.

\section{Results and Discussion}\label{sec:results}

\newlength{\figureheight}
\setlength{\figureheight}{0.54\linewidth}
\newlength{\figureheightsm}
\setlength{\figureheightsm}{0.38\linewidth}
\newlength{\figureheightmd}
\setlength{\figureheightmd}{0.48\linewidth}

\begin{figure*}[htbp]
    \centering
    % \captionsetup[subfloat]{font=scriptsize}  % Affects only this figure
    \subfloat[4\(\times\)10]{\includegraphics[height=\figureheight, trim=0 0 0 0, clip]{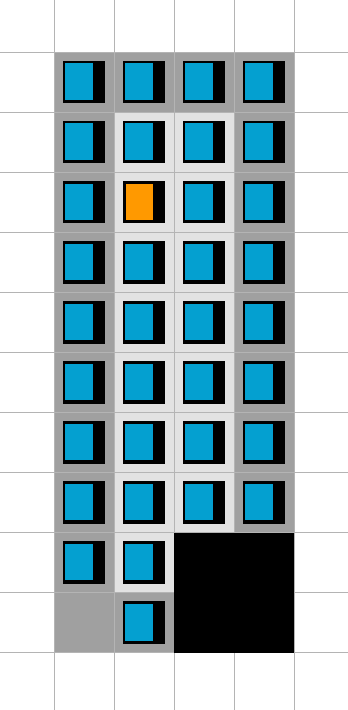}}
    \subfloat[4\(\times\)10]{\includegraphics[height=\figureheight, trim=0 0 0 0, clip]{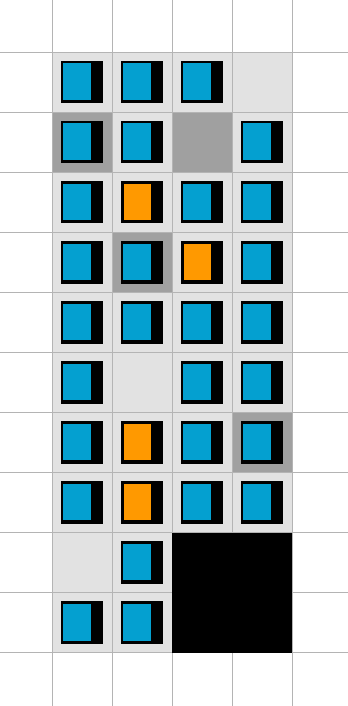}}
    \subfloat[4\(\times\)10]{\includegraphics[height=\figureheight, trim=0 0 0 0, clip]{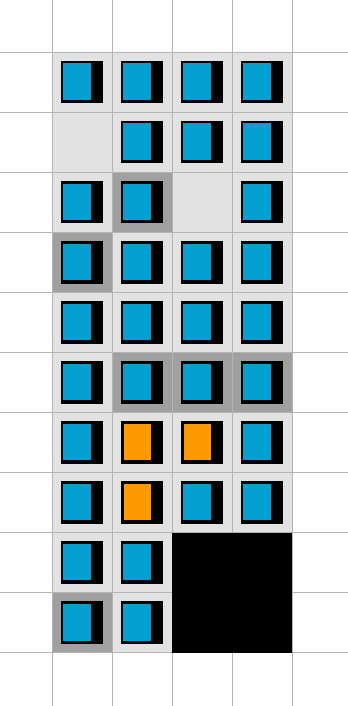}}
    \subfloat[6\(\times\)10, 6 assigned]{\includegraphics[height=\figureheight, trim=0 0 0 0, clip]{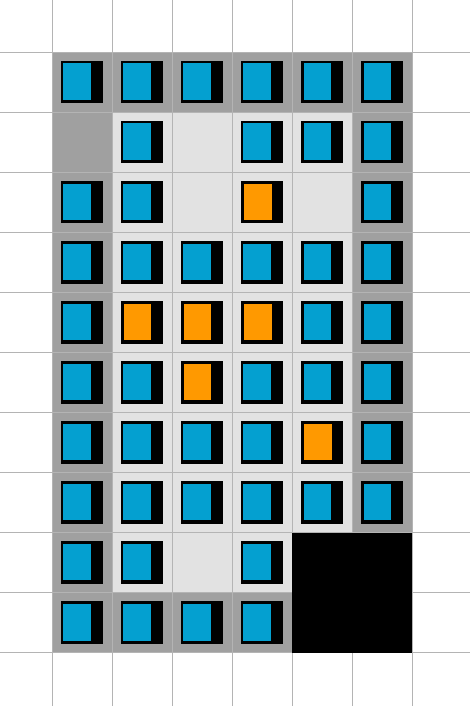}}
    \subfloat[6\(\times\)10, 5 assigned]{\includegraphics[height=\figureheight, trim=0 0 0 0, clip]{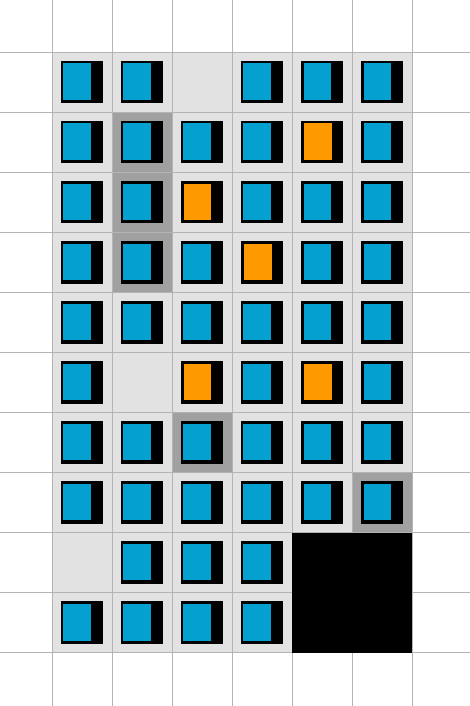}}
    \subfloat[6\(\times\)10, 5 assigned]{\includegraphics[height=\figureheight, trim=0 0 0 0, clip]{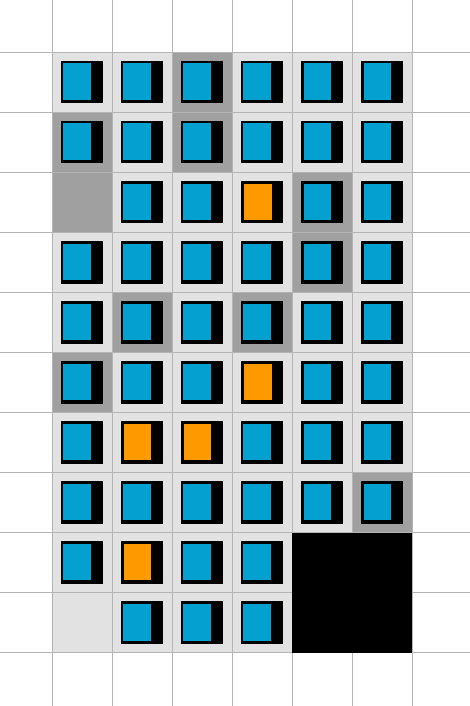}}    
    \hfill
    \subfloat[10\(\times\)10, 13 assigned,\\ 3 empties]{\includegraphics[height=\figureheightsm, trim=0 0 0 0, clip]{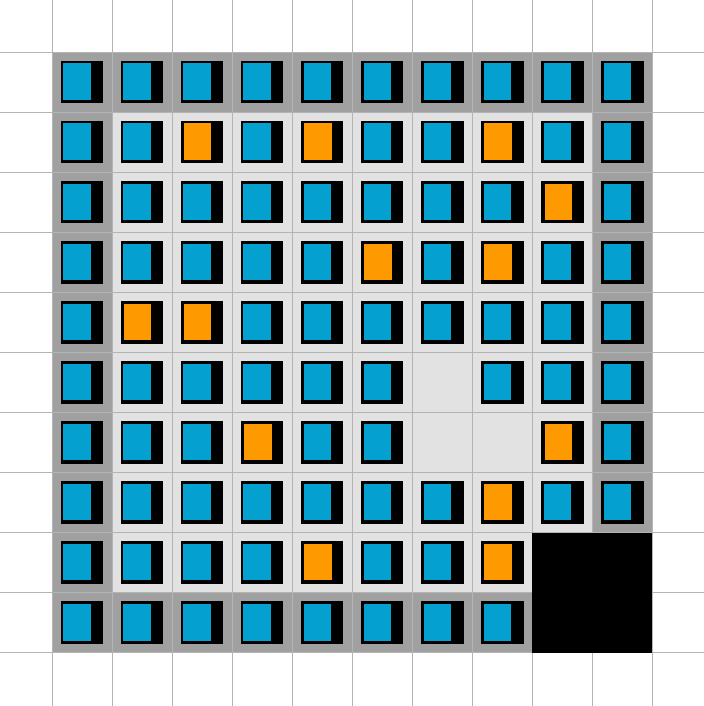}}
    \subfloat[10\(\times\)10, 1 assigned,\\ 1 empty vertex]{\includegraphics[height=\figureheightsm, trim=0 0 0 0, clip]{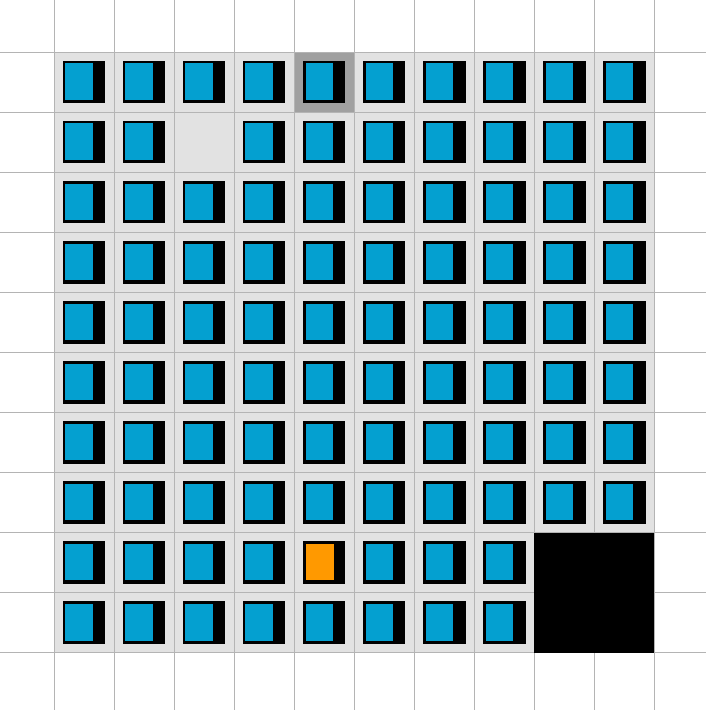}}
    \subfloat[10\(\times\)10, 13 assigned,\\ 8 empties]{\includegraphics[height=\figureheightsm, trim=0 0 0 0, clip]{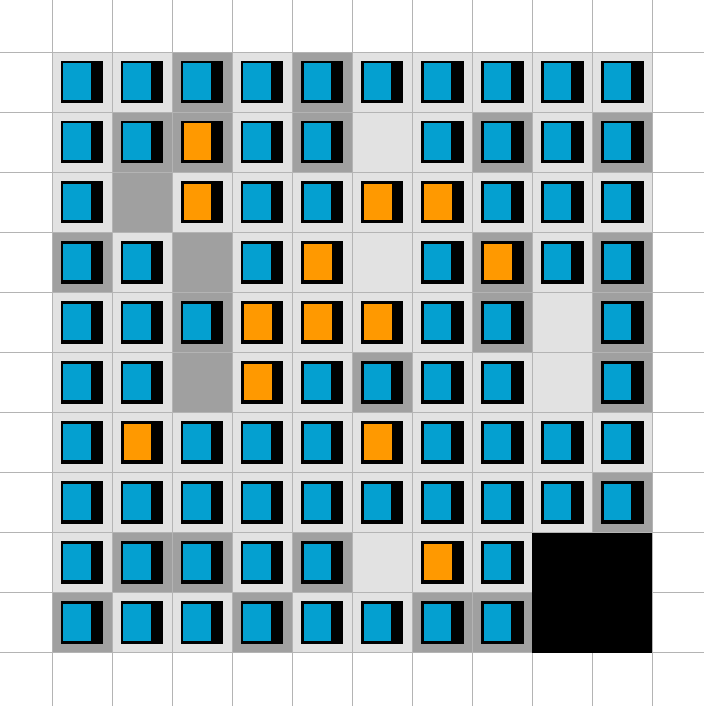}}
    \subfloat[20\(\times\)20, 40 assigned, 10 empties]{\includegraphics[height=\figureheightsm, trim=0 0 0 0, clip]{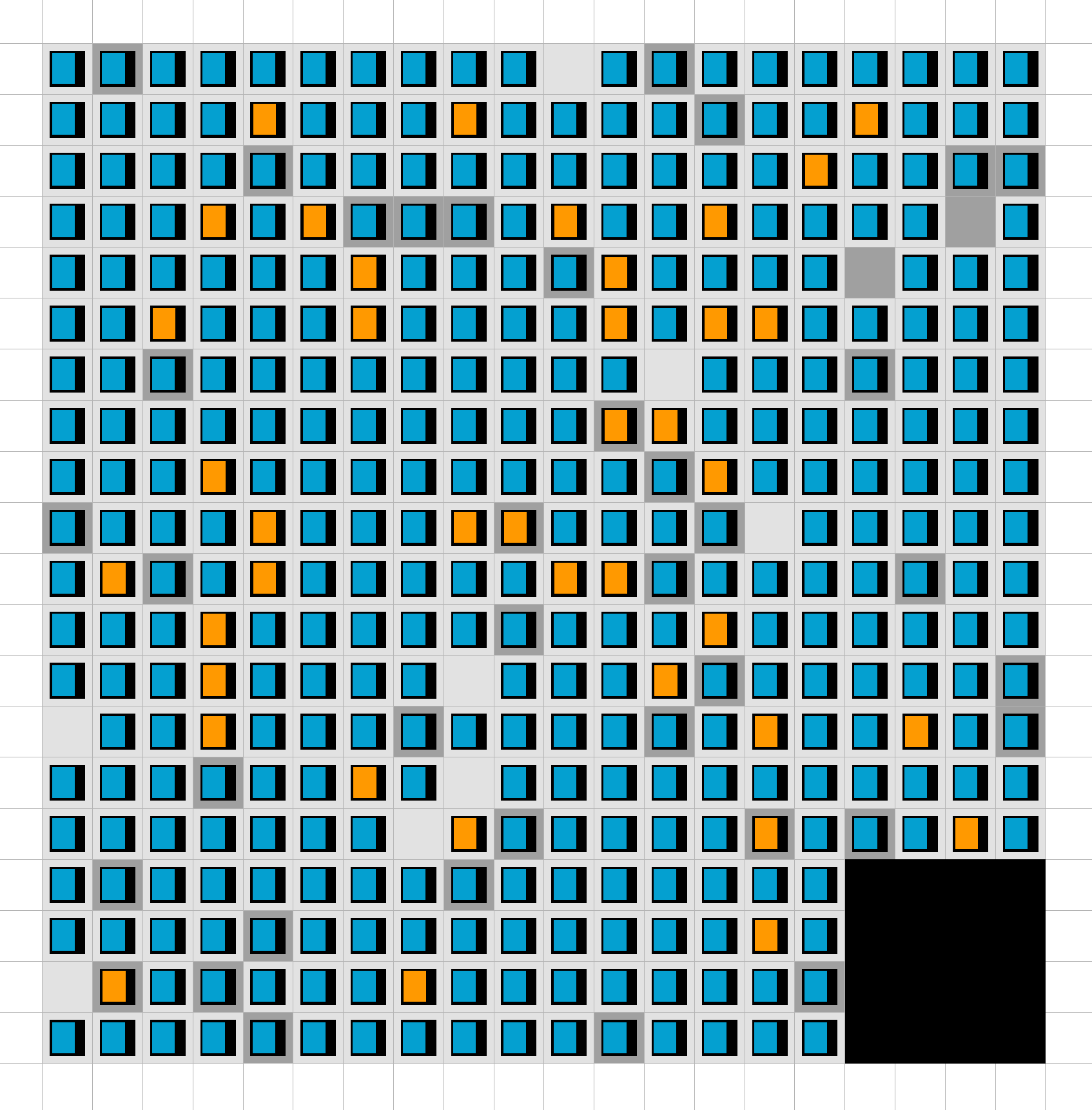}}
    \subfloat[40\(\times\)40,\\ 160 assigned, 40 empties]{\includegraphics[height=\figureheightsm, trim=0 0 0 0, clip]{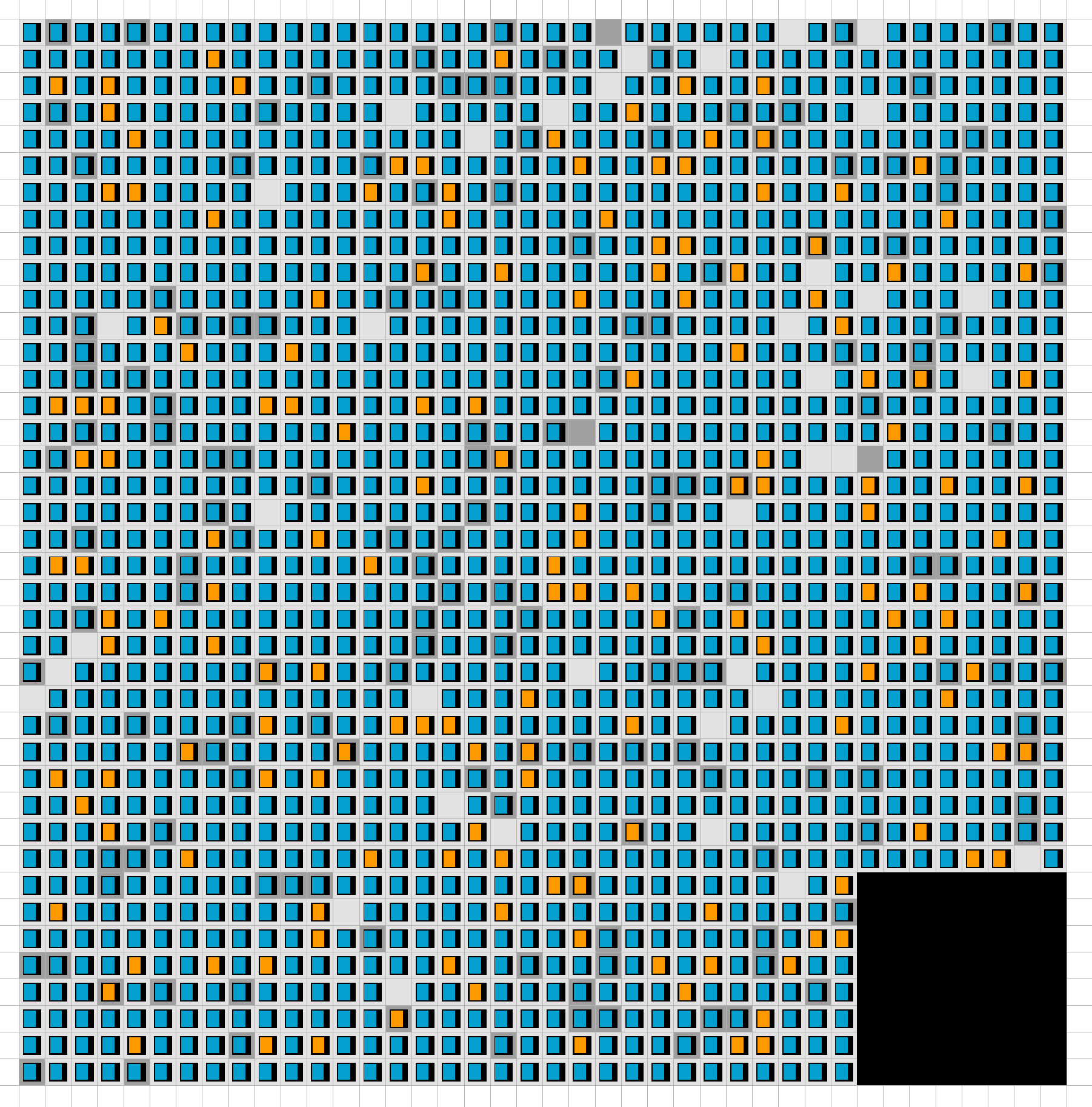}}
    \hfill 
    \subfloat[80\(\times\)80, 128 assigned,\\ 160 empties]{\includegraphics[height=\figureheightmd, trim=0 0 0 0, clip]{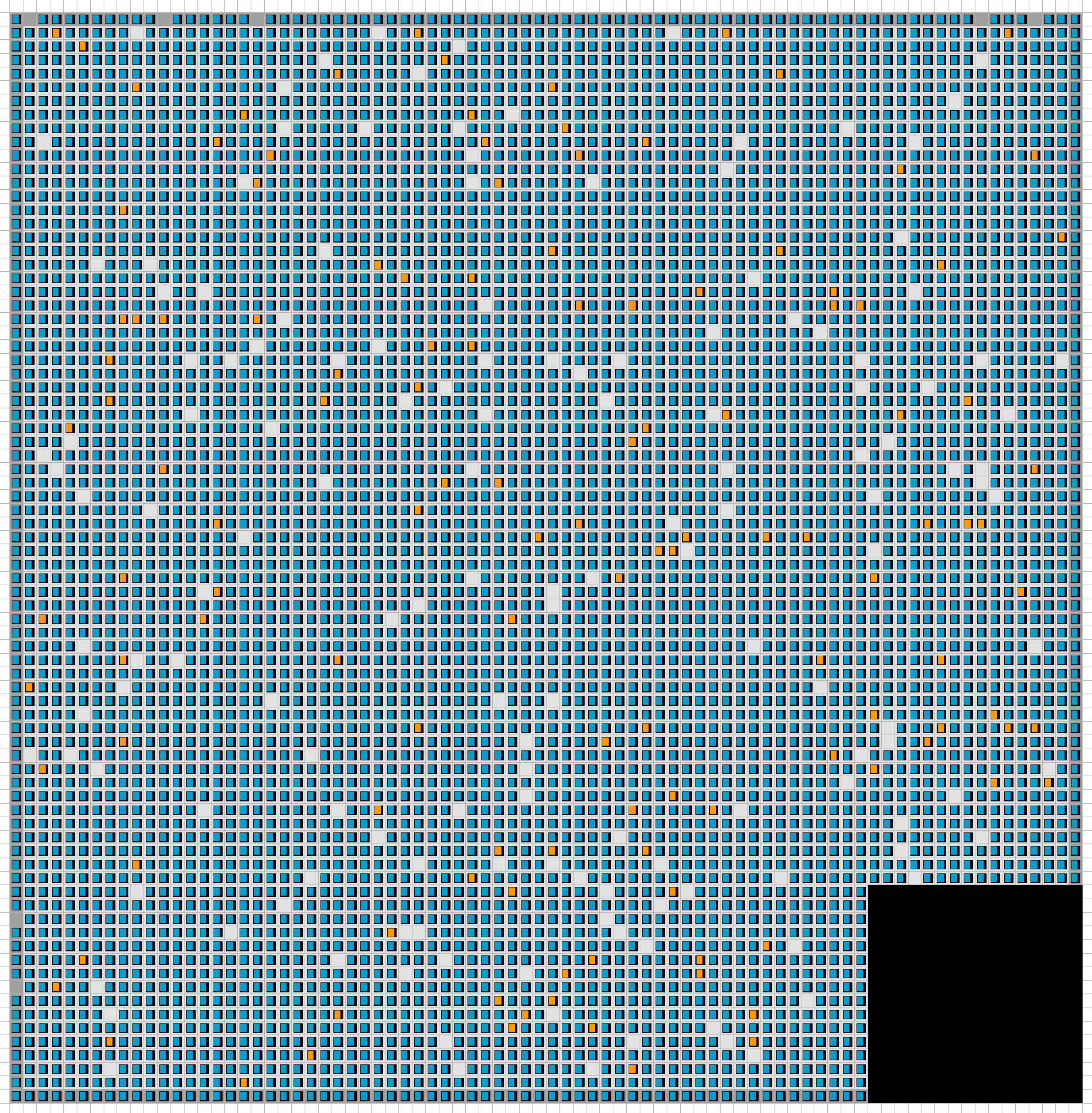}}
    \subfloat[80\(\times\)80, 128 assigned,\\ 1600 empties]{\includegraphics[height=\figureheightmd, trim=0 0 0 0, clip]{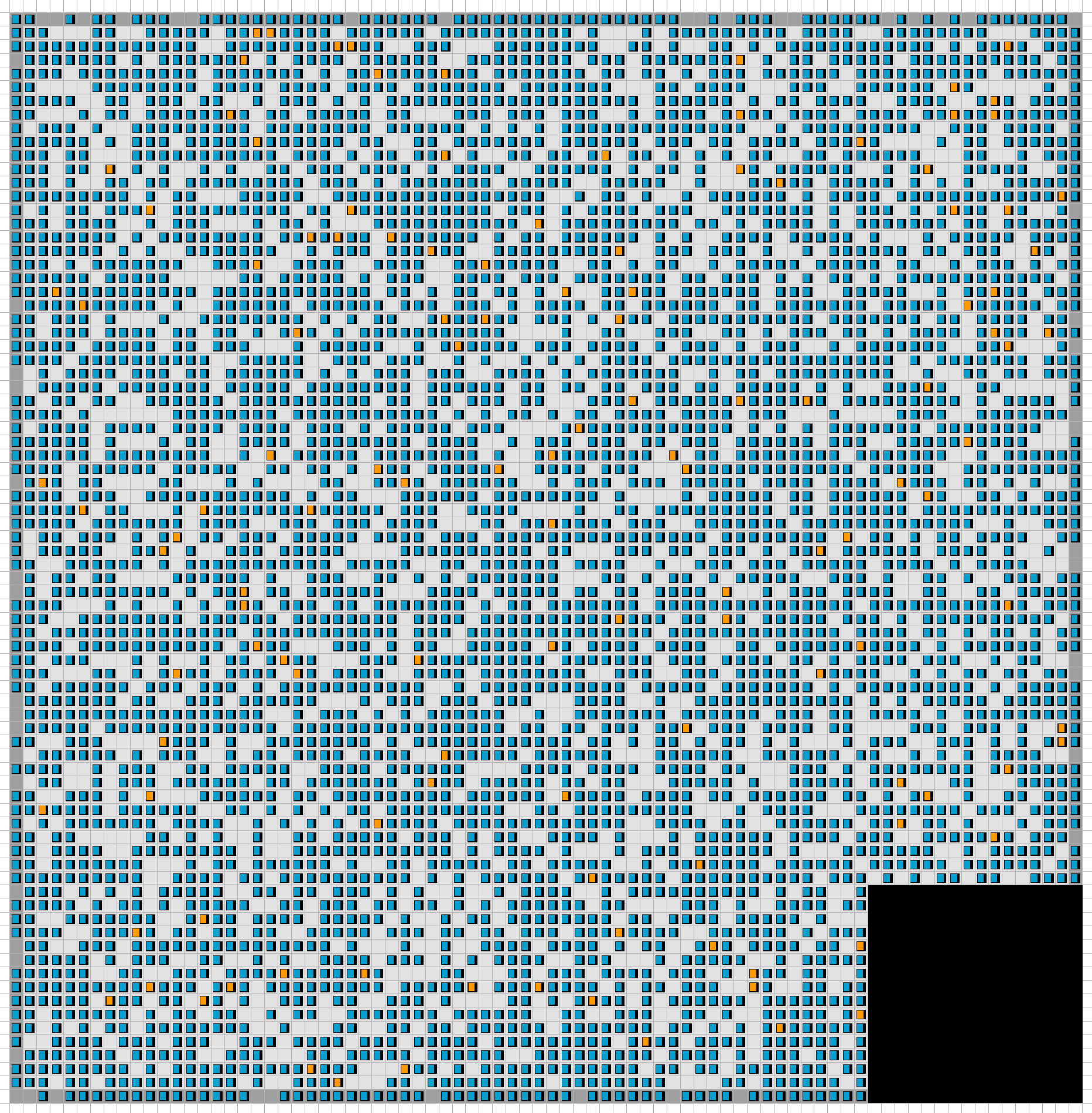}}
    \subfloat[80\(\times\)80, 800 assigned,\\ 160 empties]{\includegraphics[height=\figureheightmd, trim=0 0 0 0, clip]{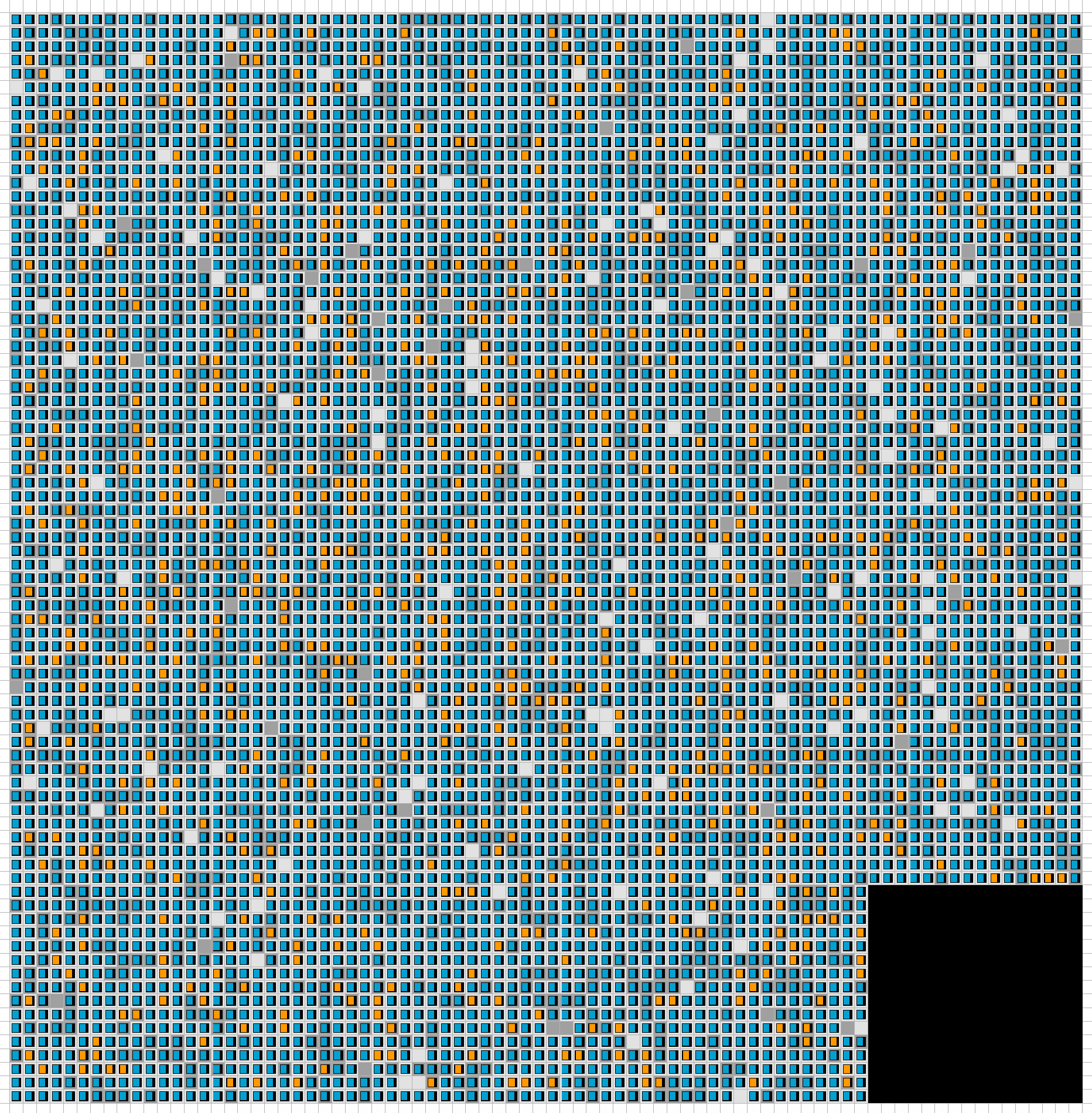}}
    \subfloat[80\(\times\)80, 800 assigned,\\ 1600 empties]{\includegraphics[height=\figureheightmd, trim=0 0 0 0, clip]{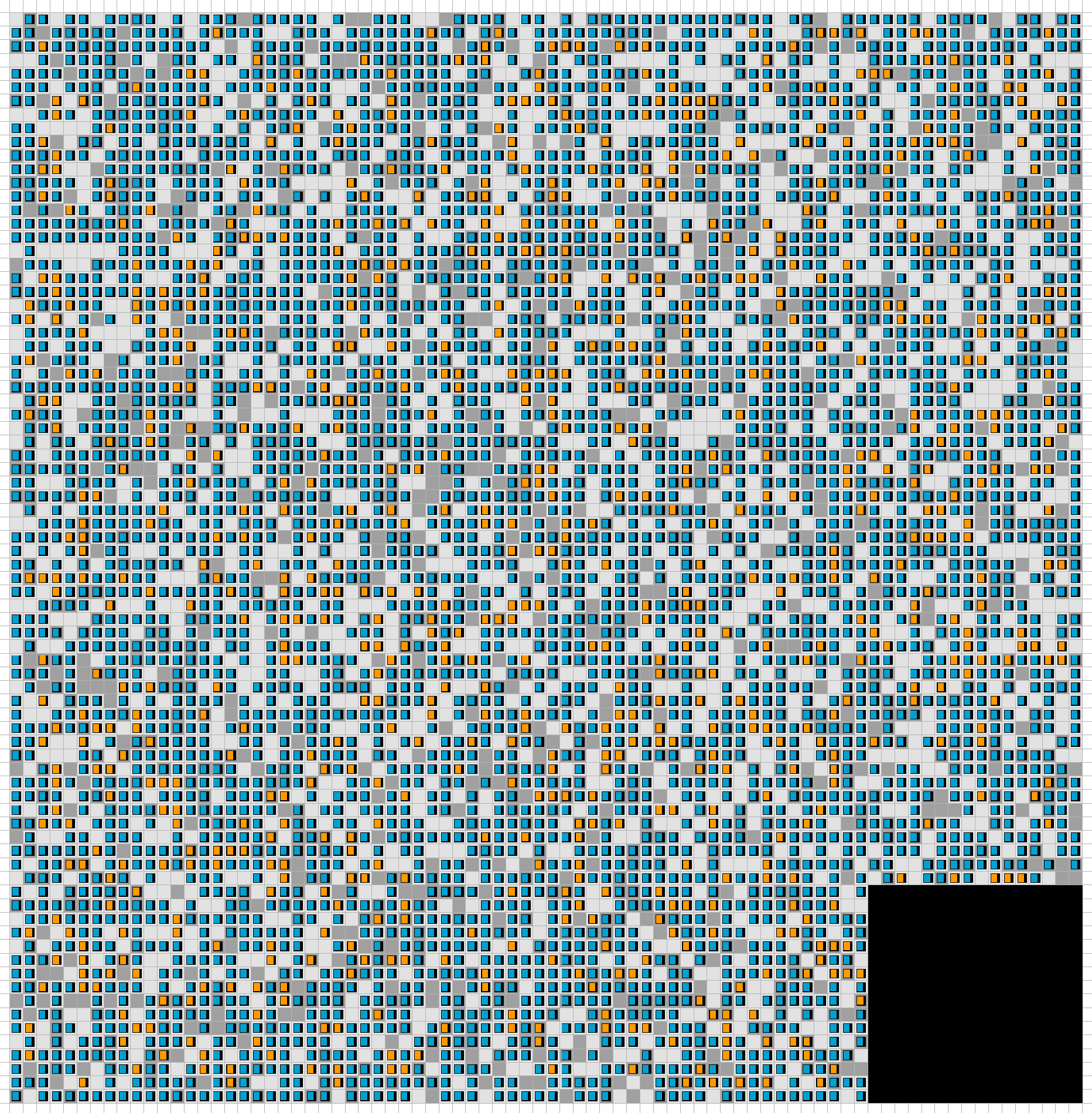}}
    \caption{Example test cases. Dark gray vertices indicate the goals for the assigned \blockL{}s. Black vertices indicate obstacles.}
    \label{fig:testcase}
\end{figure*}

\begin{table}[t]
  % \centering
  \small
    \begin{tabular}{>{\raggedright\arraybackslash}p{1.3cm}|>{\raggedright\arraybackslash}m{6.3cm}}
    \toprule
    \textbf{Parameter} & \textbf{Setup} \\
    \midrule
    \gridU{} size & 4x10, 6x10, 8x10, 10x10, 20x20, 40x40, 80x80 \\
    \hline
    Assigned \blockL{}s number & Minimum: 1\newline
    Maximum: 12.5\% of \gridL{} vertices \\
    \hline
    Empty vertex number & Minimum: 1\newline
    Maximum: 25\% of \gridL{} vertices \\
    \hline
    Goal type & Goal B: All boundary vertices\newline
    Goal R1: Random goals, 1\(\times\) \# of assigned \blockL{}s\newline
    Goal R2: Random goals, 2\(\times\) \# of assigned \blockL{}s \\
    \hline
    Random & 10 cases for each parameter combination \\
    \bottomrule
    \end{tabular}%
    \\
  \caption{Experimental setup parameters. For Goal B, the maximum assigned \blockL{} number is limited to either 12.5\% of $|V|$ or twice the \gridL{} height, whichever is smaller.}
  \label{tab:experiment-setup}%
\end{table}%

% Table generated by Excel2LaTeX from sheet 'Sheet2'
% \begin{table}[t]
%   \centering
%   \caption{Abbreviation and success rates of the algorithms.}
%     \begin{tabular}{c|l|c}
%     \toprule
%     Abbr. & \multicolumn{1}{c|}{Algorithm} & Success \\
%     \midrule
%     MAPF  & LaCAM-based Multi-agent Path Finding & 98\% \\
%     Heuristic & Heuristic Approach & 93\% \\
%     Priority & Priority-based Configuration Space Search & 48\% \\
%     Config & Configuration Space Search & 39\% \\
%     PDDL  & PDDL-based Configuration Space Search & 37\% \\
%     \bottomrule
%     \end{tabular}%
%   \label{tab:abbreviation}%
% \end{table}%

% \tabref{} \ref{tab:abbreviation} defines the abbreviations used to reference the algorithms in this section and presents their overall success rates across 13,860 test cases, where success indicates finding a feasible plan within the time limit. 

\newlength{\mycolwidth}\setlength{\mycolwidth}{0.84cm}
\newlength{\shortcolwidth}\setlength{\shortcolwidth}{0.3cm}
\begin{table*}[htbp]
% \centering
\small
\begin{tabular}{c|>{\raggedright\arraybackslash}p{\shortcolwidth}>{\centering\arraybackslash}p{\mycolwidth}>{\centering\arraybackslash}p{\mycolwidth}|>{\raggedright\arraybackslash}p{\shortcolwidth}>{\centering\arraybackslash}p{\mycolwidth}>{\centering\arraybackslash}p{\mycolwidth}|>{\raggedright\arraybackslash}p{\shortcolwidth}>{\centering\arraybackslash}p{\mycolwidth}>{\centering\arraybackslash}p{\mycolwidth}|>{\raggedright\arraybackslash}p{\shortcolwidth}>{\centering\arraybackslash}p{\mycolwidth}>{\centering\arraybackslash}p{\mycolwidth}|>{\raggedright\arraybackslash}p{\shortcolwidth}>{\centering\arraybackslash}p{\mycolwidth}>{\centering\arraybackslash}p{\mycolwidth}}
\toprule
\multirow{2}[4]{*}{Groups} & \multicolumn{3}{c}{\brlacam{}} & \multicolumn{3}{c}{Heuristic} & \multicolumn{3}{c}{Priority} & \multicolumn{3}{c}{Config} & \multicolumn{3}{c}{PDDL} \\
\cmidrule{2-16}
 & Succ. Rate & Comp. Cost & Make-span & Succ. Rate & Comp. Cost & Make-span & Succ. Rate & Comp. Cost & Make-span & Succ. Rate & Comp. Cost & Make-span & Succ. Rate & Comp. Cost & Make-span \\
\midrule
4$\times$10 & \textbf{100} & \textbf{1.00}$_{0.03}$ & \textbf{1.01}$_{0.04}$ & 93 & 1.22$_{0.31}$ & 1.34$_{0.59}$ & 91 & 1.27$_{0.31}$ & 1.53$_{0.61}$ & 87 & 1.46$_{0.38}$ & 1.77$_{0.76}$ & 89 & 1.39$_{0.46}$ & 1.72$_{0.85}$ \\
6$\times$10 & \textbf{100} & \textbf{1.01}$_{0.04}$ & \textbf{1.00}$_{0.03}$ & 97 & 1.22$_{0.26}$ & 1.40$_{0.54}$ & 87 & 1.27$_{0.26}$ & 1.53$_{0.55}$ & 76 & 1.48$_{0.41}$ & 1.82$_{0.87}$ & 76 & 1.55$_{0.73}$ & 1.96$_{1.33}$ \\
8$\times$10 & \textbf{100} & \textbf{1.01}$_{0.03}$ & \textbf{1.00}$_{0.03}$ & 97 & 1.23$_{0.24}$ & 1.45$_{0.57}$ & 75 & 1.27$_{0.23}$ & 1.54$_{0.55}$ & 60 & 1.53$_{0.48}$ & 1.90$_{1.08}$ & 54 & 1.63$_{0.93}$ & 2.01$_{1.43}$ \\
10$\times$10 & \textbf{100} & \textbf{1.00}$_{0.03}$ & \textbf{1.00}$_{0.02}$ & 96 & 1.22$_{0.23}$ & 1.50$_{0.63}$ & 60 & 1.26$_{0.21}$ & 1.56$_{0.57}$ & 42 & 1.56$_{0.60}$ & 1.96$_{1.38}$ & 34 & 1.70$_{1.27}$ & 2.04$_{1.67}$ \\
20$\times$20 & \textbf{100} & \textbf{1.01}$_{0.10}$ & \textbf{1.00}$_{0.02}$ & 95 & 1.19$_{0.17}$ & 1.67$_{0.69}$ & 15 & N/A & N/A & 7 & N/A & N/A & 4 & N/A & N/A \\
40$\times$40 & \textbf{99} & \textbf{1.04}$_{0.51}$ & \textbf{1.01}$_{0.15}$ & 91 & 1.11$_{0.12}$ & 1.56$_{0.55}$ & 8 & N/A & N/A & 3 & N/A & N/A & 0 & N/A & N/A \\
80$\times$80 & \textbf{92} & \textbf{1.04}$_{0.12}$ & \textbf{1.01}$_{0.04}$ & 80 & 1.06$_{0.09}$ & 1.34$_{0.38}$ & 2 & N/A & N/A & 0 & N/A & N/A & 0 & N/A & N/A \\
\midrule
Goal B & \textbf{100} & \textbf{1.01}$_{0.03}$ & \textbf{1.00}$_{0.03}$ & 97 & 1.13$_{0.14}$ & 1.37$_{0.44}$ & 56 & 1.20$_{0.21}$ & 1.45$_{0.46}$ & 49 & 1.49$_{0.47}$ & 1.94$_{1.05}$ & 45 & 1.51$_{0.88}$ & 1.89$_{1.28}$ \\
Goal R1 & \textbf{98} & \textbf{1.02}$_{0.34}$ & \textbf{1.01}$_{0.10}$ & 82 & 1.27$_{0.29}$ & 1.63$_{0.74}$ & 33 & 1.29$_{0.28}$ & 1.51$_{0.64}$ & 22 & 1.35$_{0.33}$ & 1.46$_{0.62}$ & 22 & 1.45$_{0.82}$ & 1.71$_{1.16}$ \\
Goal R2 & 99 & \textbf{1.02}$_{0.10}$ & \textbf{1.01}$_{0.04}$ & \textbf{100} & 1.16$_{0.22}$ & 1.42$_{0.52}$ & 56 & 1.29$_{0.27}$ & 1.56$_{0.61}$ & 47 & 1.53$_{0.48}$ & 1.85$_{1.02}$ & 44 & 1.57$_{0.76}$ & 1.96$_{1.35}$ \\
\midrule
All & \textbf{99} & \textbf{1.02}$_{0.21}$ & \textbf{1.01}$_{0.06}$ & 93 & 1.18$_{0.23}$ & 1.46$_{0.59}$ & 48 & 1.26$_{0.26}$ & 1.51$_{0.57}$ & 39 & 1.48$_{0.46}$ & 1.81$_{1.00}$ & 37 & 1.52$_{0.82}$ & 1.88$_{1.29}$ \\
\bottomrule
\end{tabular}
\caption{Algorithm performance across different configurations grouped by grid sizes and goal types, including
success rate, mean composite cost ratio, and mean makespan ratio. The subscript values are standard deviations. Smaller ratios indicate better performance. Bold values indicate the best performance across all algorithms. Ratio metrics are omitted for success rates below 20\%.
    The algorithm abbreviations are as follows.
    \brlacam{}: \lacam{} with \brpibt{},
    Heuristic: heuristic approach, 
    Priority: priority-based configuration space search,
    Config: configuration space search,
    PDDL: PDDL-based configuration space search.}
\label{tab:algorithm-performance}
\end{table*}

In this section, we evaluate the solution quality, computational time, and scalability of the algorithms proposed in Sec. \ref{sec:methods} using an extensive set of 13,860 test cases. We systematically vary \gridL{} size, number of assigned \blockL{}s, number of empty vertices, and goal location selection methods. \tabref{} \ref{tab:experiment-setup} lists the parameters used for generating these test cases, with representative examples visualized in \figref{} \ref{fig:testcase}.
Goal type B represents boundary vertex goals, while types R1 and R2 denote randomly selected goals, with type R1 being more restrictive in selection. Obstacles are modeled as a square region occupying 1/5 of the \gridL{} length, positioned in the bottom right corner of the \gridL{}.
The PDDL formulation is solved using \texttt{fast-downward}\footnote{Fast-downward: \url{https://fast-downward.org}}. All algorithms are evaluated on a single machine with AMD EPYC 7R13 Processor. The time limit for one test case is 10 seconds, in line with the typical convergence time of the algorithms.
% Or \cite{helmert2006fast}

For each test case, success indicates finding a feasible plan within the time limit.
To compare solution quality across algorithms, we establish a baseline for each test instance using the algorithm that finds the lowest final composite cost. Each algorithm's final composite cost is divided by the baseline to calculate cost ratios (resulting in ratios always greater than 1).
Makespan ratios are calculated in a similar way.

\tabref{} \ref{tab:algorithm-performance} summarizes the success rate, composite cost ratio, and makespan ratio of the proposed algorithms across different \gridL{} configurations (13,860 test cases). The \brlacam{} achieves the highest success rate, followed by the heuristic, priority, config, and PDDL algorithms. Only \brlacam{} and heuristic algorithms successfully generate solutions for large instances.
Overall, \brlacam{} and heuristic algorithms achieve the lowest composite cost and makespan ratios, followed by priority, config, and PDDL approaches. 
Note that, the \brlacam{} algorithm exhibits a higher standard deviation in composite cost on \gridL{} sizes 20$\times$20 to 80$\times$80 and for goal type R1. This variability indicates that for instances with goal vertices sparsely distributed and on large maps, \brpibt{}'s myopic behavior may leads to substantially higher solution costs than its average performance.

\begin{figure*}[htbp]
    \centering
    \includegraphics[width=\linewidth]{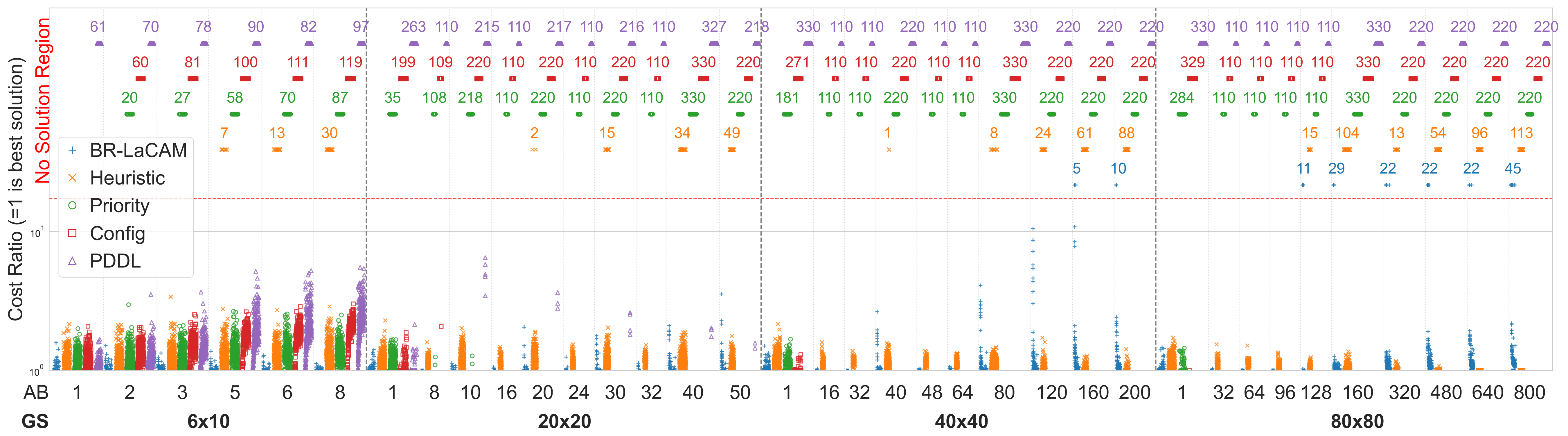}
    \caption{Composite Cost Ratio for each instance, grouped by number of assigned \blockL{}s (AB) and grid size (GS). The number and dots in the no solution region shows how many instances are unsolved for each algorithm in each group.}
    \label{fig:final_cost_ratio_by_params_dodged_all_goals}
\end{figure*}

As shown in \tabref{} \ref{tab:algorithm-performance}, algorithm success rates decrease with increasing \gridL{} size. Only the \brlacam{} and heuristic algorithms successfully solve large instances (\gridL{} size \(\geq\) 20\(\times\)20). It is important to note that the maximum number of assigned \blockL{}s increases proportionally with \gridL{} size.
Notably, \brlacam{} failures occur only when the \gridL{} size or assigned \blockL{} number exceeds computational capacity within the 10-second time limit, confirming its completeness as an algorithm - it fails only due to time constraints rather than algorithmic limitations.
Not reflected in the table, increasing the number of empty vertices reduces search depth, empirically improving success rates and reducing both composite cost and makespan.
\tabref{} \ref{tab:algorithm-performance} also demonstrates that algorithms achieve higher success rates with goal types B and R2. Goal type R1, due to its more restrictive selection criteria, results in relatively lower success rates.
\figref{} \ref{fig:final_cost_ratio_by_params_dodged_all_goals} visualizes the composite cost ratios across all 13,860 test instances, with each instance represented by a single dot.
Consistent with the results in \tabref{} \ref{tab:algorithm-performance}, failure rates increase with \gridL{} size and number of assigned \blockL{}s. The Cost Ratios among algorithms typically range between 1 and 10 relative to the best performer.

\begin{figure}[t]
    \centering
    \includegraphics[width=\linewidth]{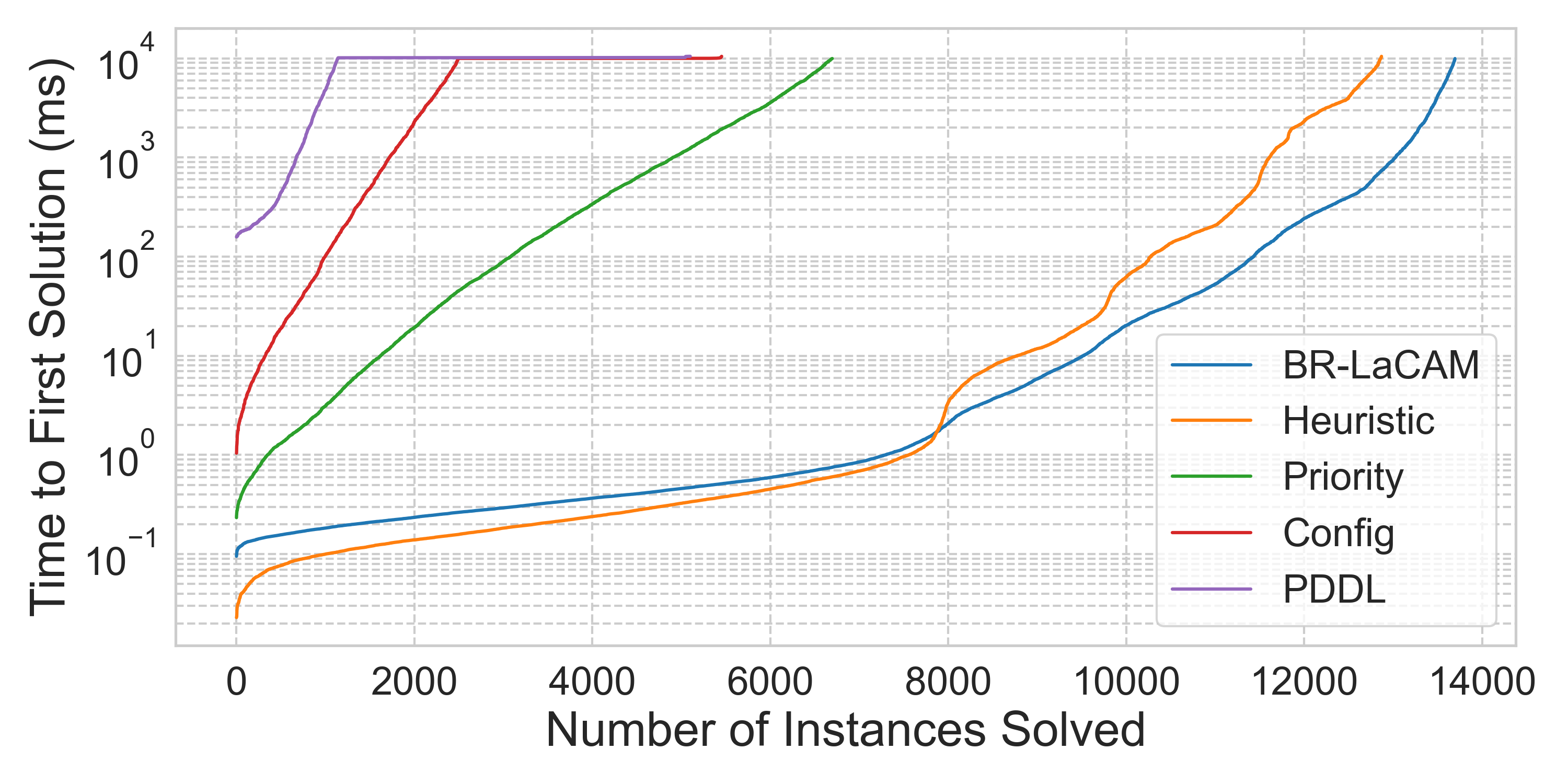}
    \caption{Time to find the first solution.}
    \label{fig:first_sol_time_cactus_all_goals}
\end{figure}

\figref{} \ref{fig:first_sol_time_cactus_all_goals} illustrates the time required for each algorithm to report its first solution. The \brlacam{} and heuristic-based algorithms find solutions within 1 millisecond for over 50\% of the test cases. \brlacam{}'s initial solution time exceeds 1 second in fewer than 10\% of cases.

\section{Conclusion}\label{sec:conclusions}

This paper formally defines the \problemupperP{} and establishes its relationship to sliding puzzles. We develop five solution algorithms: configuration space search, PDDL-based planning, priority-based search, \brlacam{}, and heuristic-based methods.
Through extensive evaluation across 13,860 test cases with varying \gridL{} sizes, assigned \blockL{} numbers, and goal configurations, we evaluate the effectiveness of the proposed approaches and provide insights into their relative strengths.
While most algorithms show degraded success rates with increasing \gridL{} size and more restrictive goal choices, the \brlacam{} and heuristic-based approaches successfully solve large instances, achieving 99\% and 93\% success rates respectively with dense \gridL{}s from 4\(\times\)10 to 80\(\times\)80. Among the proposed methods, the \brlacam{} and heuristic approaches demonstrate superior performance in both solution quality (composite cost and makespan) and computational efficiency, finding solutions within 1 millisecond for over 50\% of test cases and within 1 second for over 90\% of cases. These results establish a comprehensive baseline for solving \problemupperP{}s.

Based on our comprehensive analysis, future research directions include exploring various problem variants including different terminal states (where assigned \blockL{}s become empty vertices, obstacles, or unassigned blocks), additional objective functions, constraints on simultaneous actions, and practical priority requirements. Given \brlacam{} approach's superior performance in this work, improving its solution quality and scalability through enhanced goal selection processes and 
informative action preference heuristics
% BRaP-specific node expansion heuristics 
emerges as a particularly promising direction.

\appendix
\section*{Appendix: Admissible Heuristic for Configuration Space Search}\label{appendix:heuristic}

\begin{figure}[h!]
    \centering
    \includegraphics[width=0.65\linewidth]{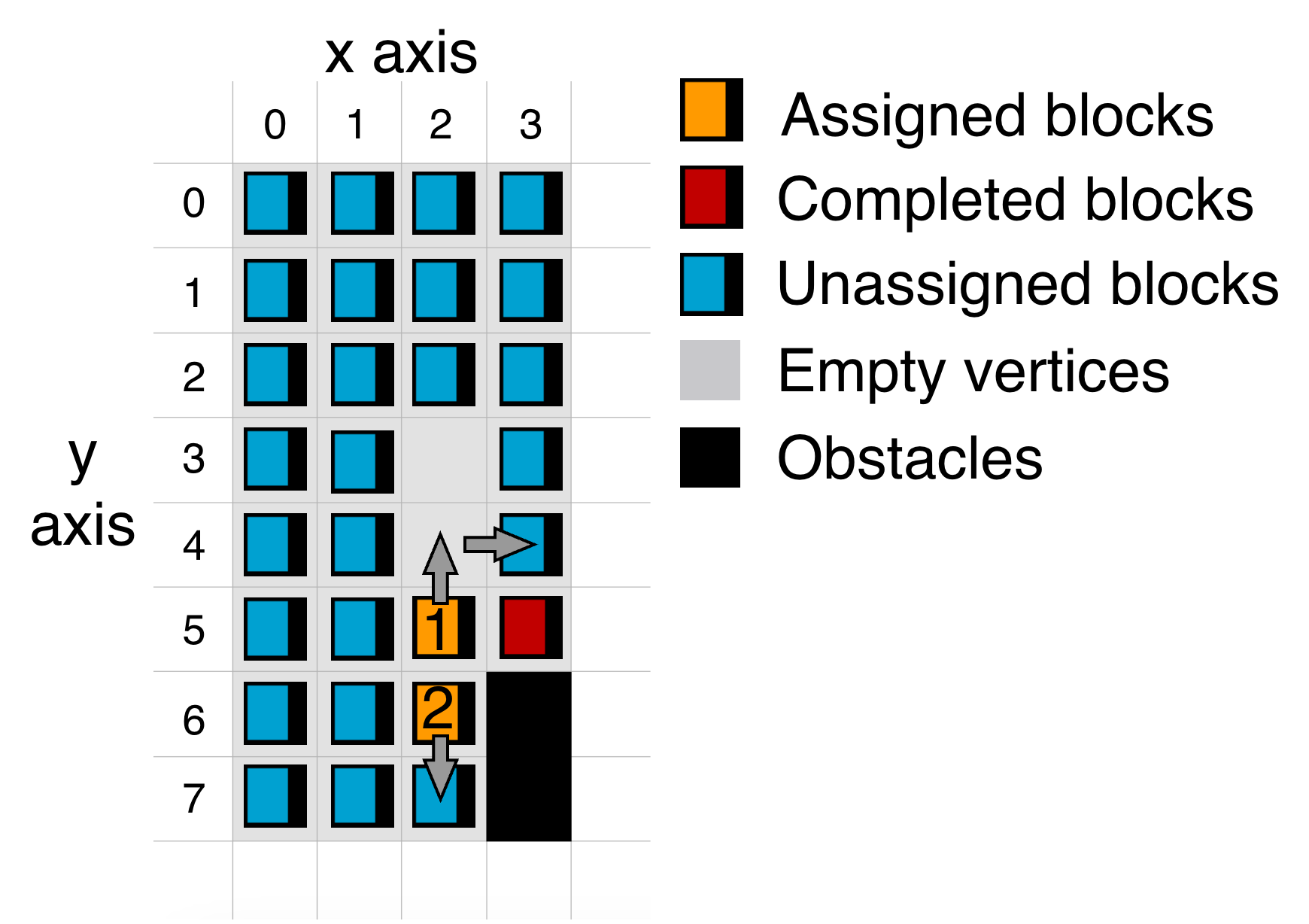}
    \caption{Illustration of the formulated heuristic. The goal of the assigned \blockL{}s is the boundary.}
    \label{fig:multipod-heuristic}
\end{figure}

An admissible heuristic is formulated by assuming assigned \blockL{}s move along least-blocking vertices and each blocking \blockL{} can be cleared using only one \move{} action.
Using the example state \(s\) in \figref{} \ref{fig:multipod-heuristic}, we demonstrate the heuristic value calculation for two assigned \blockL{}s. We calculate a lower bound cost for moving each assigned \blockL{} to its goal set.
\blockU{} 1 requires: an initial \move{}, clearing the \blockL{} at (3, 4) (minimum one \move{}), a final \move{}, and a \complete{} action. Therefore, the lower bound cost for \blockL{} 1 is $h(s, 1) = c_{\move{}} + c_{\move{}} + c_{\move{}} + c_{\complete{}} = 3 c_{\move{}} + c_{\complete{}}$.
Similarly, the movement of \blockL{} 2 to the goals requires two \move{} actions and one \complete{} action, yielding a lower bound cost of $h(s, 2) = c_{\move{}} + c_{\move{}} + c_{\complete{}} = 2 c_{\move{}} + c_{\complete{}}$.
Since actions moving one assigned \blockL{} can benefit the movement of other assigned \blockL{}, we estimate a joint cost lower bound by averaging individual costs. Thus, for this example:
\(h(s) = (h(s,1) + h(s,2)) / 2 = [(3 c_{\move{}} + c_{\complete{}}) + (2 c_{\move{}} + c_{\complete{}})] / 2 = 2.5 c_{\move{}} + c_{\complete{}}\).

Let the lowest action cost to move the assigned \blockL{}s to the goals be \(h^*(s)\).

\textbf{Claim 2}: \(h(s)\) is admissible: \(h(s) \leq h^*(s) \).
\begin{proof}
Let the lowest cost to move each assigned \blockL{} \(i \in \Ical\) to their goal vertices be \(h^*(s, i)\). By definition, \(h(s, i)\) underestimates the cost to move \blockL{} \(i\). Therefore, \(h(s, i) \leq h^*(s, i) \leq h^*(s)\). Therefore, \(h(s) = \frac{1}{|\Ical|} \sum_{i \in \Ical} h(s, i) \leq \frac{1}{|\Ical|} |\Ical| h^*(s) = h^*(s)\).
\end{proof}

% \textbf{Claim 2}: \(h(s)\) is consistent: \(h(s) \leq c(s, s') + h(s')\).

% \input{section/appendix/pddl_domain}
% \input{section/appendix/additional_results}

\bibliography{ms}

% \bibliography{aaai2026}

\end{document}